\documentclass[journal]{IEEEtran}
\usepackage[T1]{fontenc}
\ifCLASSINFOpdf
\else
\fi

\usepackage{cuted}
\usepackage{graphicx}
\usepackage{amsfonts}
\usepackage{amsmath}
\usepackage[colorlinks,linkcolor=blue,anchorcolor=blue,citecolor=blue]{hyperref}
\usepackage{listings}
\usepackage{amssymb}
\usepackage{amsthm}
\usepackage{graphicx}
\usepackage{epstopdf}
\usepackage{setspace}
\usepackage{booktabs}
\usepackage{threeparttable}
\usepackage{multirow}
\usepackage[mathlines]{lineno}
\usepackage{subfigure}
\usepackage{makecell}
\usepackage[linesnumbered,ruled,vlined,algo2e]{algorithm2e}
\usepackage{threeparttable}
\usepackage{mathrsfs}
\theoremstyle{plain}

\theoremstyle{definition}
  \newtheorem{defn}{Definition}

  \newtheorem{rem}{Remark}

\theoremstyle{break}

\theoremstyle{definition}

\begin{document}
%
% paper title
% Titles are generally capitalized except for words such as a, an, and, as,
% at, but, by, for, in, nor, of, on, or, the, to and up, which are usually
% not capitalized unless they are the first or last word of the title.
% Linebreaks \\ can be used within to get better formatting as desired.
% Do not put math or special symbols in the title.
\title{\huge LGBQPC: Local Granular-Ball Quality Peaks Clustering}
%
%
% author names and IEEE memberships
% note positions of commas and nonbreaking spaces ( ~ ) LaTeX will not break
% a structure at a ~ so this keeps an author's name from being broken across
% two lines.
% use \thanks{} to gain access to the first footnote area
% a separate \thanks must be used for each paragraph as LaTeX2e's \thanks
% was not built to handle multiple paragraphs
%

\author{Zihang~Jia, Zhen~Zhang, \IEEEmembership{Senior Member, IEEE}
and Witold Pedrycz, \IEEEmembership{Life Fellow, IEEE}
\thanks{This work was partly supported by the National Natural Science Foundation of China under Grant Nos. 72371049 and 71971039, the Natural Science Foundation of Liaoning Province under Grant No. 2024-MSBA-26 and the Funds for Humanities and Social Sciences of Ministry of Education of China under Grant No. 23YJC630219.
(\emph{Corresponding author: Zhen Zhang.})}
\thanks{Zihang Jia and Zhen Zhang are with the Institute of Systems Engineering, School of Economics and Management, 
Dalian University of Technology, Dalian 116024, China (e-mails:
zihangjia@outlook.com; zhen.zhang@dlut.edu.cn).
}% <-this % stops a space
\thanks{Witold Pedrycz is with the Department of Measurement and Control Systems, Silesian University of Technology (SUT), 44-100 Gliwice, Poland, also with the Department of Electrical and Computer Engineering, University of Alberta, Edmonton, AB T6G 2R3, Canada, also with the Institute of Systems Engineering, Macau University of Science and Technology, Macau, China, and also with the Research Center of Performance and Productivity Analysis, Istinye University, 34010 Istanbul, Türkiye (e-mail: wpedrycz@ ualberta.ca).
}
}

\maketitle

\begin{abstract}
  The density peaks clustering (DPC) algorithm has attracted considerable attention for its ability to detect arbitrarily shaped clusters based on a simple yet effective assumption. Recent advancements integrating granular-ball (GB) computing with DPC have led to the GB-based DPC (GBDPC) algorithm, which improves computational efficiency. However, GBDPC demonstrates limitations when handling complex clustering tasks, particularly those involving data with complex manifold structures or non-uniform density distributions. To overcome these challenges, this paper proposes the local GB quality peaks clustering (LGBQPC) algorithm, which offers comprehensive improvements to GBDPC in both GB generation and clustering processes based on the principle of justifiable granularity (POJG). Firstly, an improved GB generation method, termed GB-POJG+, is developed, which systematically refines the original GB-POJG in four key aspects: the objective function, termination criterion for GB division,  definition of abnormal GB, and granularity level adaptation strategy. GB-POJG+ simplifies parameter configuration by requiring only a single penalty coefficient and ensures high-quality GB generation while maintaining the number of generated GBs within an acceptable range. In the clustering phase, two key innovations are introduced based on the GB $k$-nearest neighbor graph: relative GB quality for density estimation and geodesic distance for GB distance metric. These modifications substantially improve the performance of GBDPC on datasets with complex manifold structures or non-uniform density distributions. Extensive numerical experiments on 40 benchmark datasets, including both synthetic and publicly available datasets, validate the superior performance of the proposed LGBQPC algorithm.
\end{abstract}


% For peer review papers, you can put extra information on the cover
% page as needed:
% \ifCLASSOPTIONpeerreview
% \begin{center} \bfseries EDICS Category: 3-BBND \end{center}
% \fi
%
% For peerreview papers, this IEEEtran command inserts a page break and
% creates the second title. It will be ignored for other modes.
\IEEEpeerreviewmaketitle

\section{Introduction}\label{Section1}

\IEEEPARstart{C}{lustering} analysis is a fundamental task in unsupervised machine learning \cite{XuRui2005TNNLS}, aiming to uncover inherent patterns and structures within data distributions without relying on instance labels. In this context, a cluster represents a group of instances exhibiting higher similarity to each other than to those in other clusters. Over the past few decades, numerous clustering algorithms have been developed to tackle diverse  tasks, such as $k$-means clustering \cite{Arthur2007ACM-SIAM}, spectral clustering (SC) \cite{VonLuxburg2007StatisticsComputing},  density-based spatial clustering of applications with noise (DBSCAN) \cite{Ester1996KDD}, and density peaks clustering (DPC) \cite{Rodriguez2014Science}.

Among these algorithms, DPC has attracted considerable attention due to its straightforward assumptions and  capability to identify clusters with arbitrary shapes \cite{WangYizhang2024ESWA}. Specifically, DPC operates on two fundamental premises: (1) a cluster center exhibits a higher density than its neighbors, and (2) it maintains a relatively large distance from any instance with a higher density. Following the identification of cluster centers, non-center instances are assigned to cluster centers according to their nearest neighbor with higher density. Nevertheless, DPC has a time and space complexity of $O(n^{2})$, where $n$ denotes the number of instances, thereby limiting its applicability to large datasets. Research efforts have been directed toward addressing various challenges associated with DPC, such as clustering data with non-uniform density \cite{HouJian2020TII}, handling large-scale data \cite{FangXintong2023TII}, clustering imbalanced data \cite{TongWuning2023TKDE}, developing parameter-free algorithms \cite{dErrico2021INS}, clustering data with complex manifold structures \cite{DuMingjing2018IJMLC}, and clustering heterogeneous data \cite{DingShifei2017KBS}.

Recently, to enhance the efficiency and robustness of clustering results, Cheng et al. \cite{ChengDongdong2024TNNLS} developed an improved DPC algorithm by integrating granular-ball (GB) computing, resulting in the GB-based DPC (GBDPC) algorithm. GB computing, initially introduced by Xia et al. \cite{XiaShuyin2019INS}, is an efficient, robust, and interpretable computing paradigm. Within this paradigm, GBs with multiple granularities are efficiently generated from instances with single granularity and serve as the fundamental computing units. Notably, the number of generated GBs is considerably lower than that of the original instances, thereby significantly enhancing computational efficiency. Consequently, the GBDPC algorithm necessitates only the calculation of the density of each GB and the distances between GBs, without depending on the finest-grained instances, thus achieving robust clustering results efficiently. Based on the GB computing paradigm, several GB-based clustering algorithms have been developed, including GB-based spectral clustering (GBSC) \cite{XieJiang2023TKDE}, GB-based DBSCAN (GB-DBSCAN) \cite{ChengDongdong2024INS},  GBCT \cite{XiaShuyin2024TNNLS}, GB-FuzzyStream for stream clustering \cite{XieJiang2024ICDE}, and W-GBC for high-dimensional data clustering \cite{XieJiang2024ICDE2}.

However, GBDPC demonstrates limitations in effectively handling complex clustering tasks, particularly those involving datasets  with non-uniform density distribution or complex manifolds. These limitations can be attributed to the following factors:
\begin{enumerate}
  \item [(1)] The performance of GBDPC heavily relies on the quality of the generated GBs. To address this, based on the principle of justifiable granularity (POJG), Jia et al. \cite{JiaZihang2025TCYB} developed GB-POJG, an advanced GB generation method specifically designed for clustering tasks. While GB-POJG have significantly enhanced the performance of GBDPC, it demonstrates two inherent limitations: (\romannumeral1) it necessitates the setting of two parameters, and (\romannumeral2) it cannot generate GBs at a high granularity level while ensuring the number of generated GBs within an acceptable range.
  \item [(2)] In the process of GB clustering, GBDPC  exhibits two fundamental challenges: (\romannumeral1) its density estimation mechanism overlooks the integration of local structural information for each GB, significantly undermining its performance on datasets with non-uniform density distrbution; and (\romannumeral2) its reliance on Euclidean distance for calculating distance between GBs renders it insufficient to identify complex manifolds.
\end{enumerate}

To address these limitations, this paper aims to develop a novel GB-based clustering algorithm through improving both the GB generation and clustering processes. The main contributions of this paper are summarized as follows:
\begin{enumerate}
  \item [(1)] To address the inherent limitations of GB-POJG, this paper introduces GB-POJG+, a comprehensive refinement of GB-POJG, focusing on four aspects: objective function, termination criterion for GB division, definition of abnormal GBs, and adaptation strategy for granularity levels. To the best of our knowledge, GB-POJG+ is the only GB generation method that balances the quality and quantity of generated GBs in clustering tasks.
  \item [(2)] To effectively cluster data exhibiting non-uniform densities or complex manifolds, a $k$-nearest neighbor ($k$-NN) graph is established for the generated GBs. By leveraging the GB $k$-NN graph, the relative quality is introduced as a density estimator to sufficiently consider local structural information of GBs, while geodesic distance is utilized as the distance metric for GBs to identify complex manifolds.
  \item [(3)] Integrating the above contributions, a novel clustering algorithm, termed local GB quality peaks clustering (LGBQPC), is developed to efficiently tackle complex clustering tasks.
\end{enumerate}

The remainder of this paper is structured as follows. Section \ref{Section2} presents a thorough review of related works. Section \ref{Section3} details the LGBQPC algorithm. Section \ref{Section4} presents a series of numerical experiments to validate the efficacy of the LGBQPC algorithm. Section \ref{Section5} summarizes the paper and suggests future research directions.

\section{Related Works}\label{Section2}
This section reviews foundational concepts and related research on GB computing, GB-POJG and GBDPC. For convenience,  the notations used throughout this paper are first introduced as follows. Let $U=\left\{\mathbf{x}_{1},\cdots,\mathbf{x}_{n}\right\}\subset\mathbb{R}^{m}$ denote the universe of all instances, where $n$ and $m$ represent the number of instances and features, respectively, and $\mathbb{R}$  is the real domain. The maximum pairwise distance between instances in $U$ is denoted as $d_{\max}$. For any set $X$, its cardinality is denoted as $\vert X\vert$. Let $\mathbf{y}$ be a vector, then $\Vert \mathbf{y}\Vert_{2}$, $\text{mean}\left(\mathbf{y}\right)$, $\text{median}(\mathbf{y})$, and $\text{std}(\mathbf{y})$ represent the 2-norm, mean, median and standard deviation of the elements in $\mathbf{y}$, respectively.

\subsection{Granular-Ball Computing}\label{Section:Granular-Ball Computing}
Granular-ball computing is a computing paradigm consisting of two main steps: (1) generating GBs from the universe, and (2) learning a GB-based model. The formal definition of a GB is as follows.
\begin{defn} (See \cite{XiaShuyin2019INS,XiaShuyin2024TNNLS})
  Let $X\subseteq U$. The GB derived from $X$ is defined as a 4-tuple $\Omega_{X}=\left(X,\mathbf{c}_{X},R_{X}^{\text{ave}},R_{X}^{\text{max}}\right)$, where $\mathbf{c}_{X}=\left(\sum_{\mathbf{x}_{i}\in X}\mathbf{x}_{i}\right)/\vert X\vert$, $R_{X}^{\text{ave}}=\left(\sum_{\mathbf{x}_{i}\in X}\Vert \mathbf{x}_{i}-\mathbf{c}_{X}\Vert_{2}\right)/\vert X\vert$
  and $R_{X}^{\text{max}}=\max_{\mathbf{x}_{i}\in X}\Vert \mathbf{x}_{i}-\mathbf{c}_{X}\Vert_{2}$ are the center, average radius and maximum radius of $\Omega_{X}$, respectively. Each $\mathbf{x}_{i}\in X$ is always considered to belong to $\Omega_{X}$.
\end{defn}

Mathematically, generating GBs involves partitioning the universe into disjoint subsets $\Pi=\{X_{1},\cdots,X_{t}\}$ such that instances in each subset $X_{i}$ are as similar as possible. This yields a corresponding set of GBs, $\Phi = \{\Omega_{X_1}, \cdots, \Omega_{X_t}\}$. The learning process of a GB-based model follows the standard model of GB computing, which can be expressed as: $g(\mathbf{x},\boldsymbol{\theta})\rightarrow g^{*}(\Omega_{X},\boldsymbol{\theta}^{*})$ \cite{XieJiang2024TPAMI}. Here, $g(\textbf{x},\boldsymbol{\theta})$ refers to a traditional instance-driven learning model with a model parameter vector $\boldsymbol{\theta}$, while $g^{*}(\Omega_{X},\boldsymbol{\theta}^{*})$ represents a GB-driven learning model with a model parameter vector $\boldsymbol{\theta}^{*}$. In summary, GB computing improves the model's learning efficiency by efficiently generating GBs and using them as input instead of individual instances. Current research on GB computing primarily focuses on the efficient generation of GBs \cite{JiaZihang2025TCYB,XieQin2024IEEETETCI,XiaShuyin2024TNNLS2} and their applications to various tasks, such as classification \cite{Sajid2025PR,YangJie2024IEEETFS}, feature selection \cite{SunLin2024TFS,QianWenbin2024TKDE}, sampling \cite{XiaShuyin2023TNNLS}, clustering \cite{ChengDongdong2024TNNLS,XieJiang2023TKDE}, and outlier detection \cite{GaoCan2025TKDE,ChengShitong2024PR}.

\subsection{GB-POJG}

Generating GBs for clustering tasks is challenging due to the absence of instance labels, which makes evaluating GB quality difficult. The POJG provides a well-established framework for designing and assessing information granules based on their coverage and specificity, without relying on instance labels or any specific formal construct \cite{Pedrycz2013ASOCO,Pedrycz2024TCYB}. Building upon the POJG, the GB-POJG approach treats GBs as information granules and defines their coverage and specificity, thus providing a comprehensive assessment of GB quality. Moreover, GB-POJG generates GBs in an unsupervised manner by maximizing the overall quality of the generated GBs, ensuring their alignment with the underlying data distribution.
\begin{defn}\label{Definition:Quality of Granular-Ball} (See \cite{JiaZihang2025TCYB})
Let $f_{1}$ be an increasing function, and $f_{2}$ a decreasing function. The quality level of a GB $\Omega_{X}$ is defined as $\mathcal{Q}(\Omega_{X})=\mathcal{Q}_{C}(\Omega_{X})\cdot\mathcal{Q}_{S}(\Omega_{X})$, where $\mathcal{Q}_{C}(\Omega_{X})=f_{1}\left(\left\vert \left\{\mathbf{x}_{i}\in X:\Vert \mathbf{x}_{i}-\mathbf{c}_{X}\Vert_{2}\leq R_{X}^{\text{ave}}\right\}\right\vert\right)$ and $\mathcal{Q}_{S}(\Omega_{X})=f_{2}\left(R_{X}^{\text{ave}}\right)$ are the coverage and specificity of $\Omega_{X}$, respectively.
\end{defn}
\begin{rem}\label{Remark:Granularity-Level}
In \cite{JiaZihang2025TCYB}, the increasing function $f_{1}$ and the decreasing function $f_{2}$ are specified as $f_{1}(t)=t$ and $f_{2}(t)=\exp(-\gamma\cdot t)$, respectively, where $\gamma\geq0$ is a parameter termed granularity level. Notably, as $\gamma$ increases, the specificity becomes increasingly significant, thereby encouraging the generation of GBs with smaller average radii. Moreover, the number of GBs generated by GB-POJG typically increases with the granularity level.
\end{rem}
Based on Definition \ref{Definition:Quality of Granular-Ball}, the objective function of GB-POJG is the oveall quality of all generated GBs, which can be formally formulated as 
\begin{align}\label{Equation:Objective-Function-GB-POJG}
  J(\Phi)=\sum\nolimits_{i=1}^{t}\mathcal{Q}(\Omega_{X_{i}}),
\end{align}
where $\Phi=\{\Omega_{X_{1}},\cdots,\Omega_{X_{t}}\}$ denotes the set of generated GBs. To maximize the overall quality, GB-POJG follows a top-down paradigm \cite{XiaShuyin2019INS} and conducts a pre-division of GBs. To be specific, GB-POJG iteratively divides GBs, commencing from $\Omega_{U}$, the GB of the entire universe $U$, to ensure that any GB is derived from instances belonging to the same cluster. This process continues until all GBs are sufficiently small. Notably, the following 2-division method is adopted to divide a given GB $\Omega_{X}$ \cite{XieJiang2024TPAMI}: let  
$\mathbf{x}_{\alpha}=\arg\max_{\mathbf{x}_{i}\in X}\Vert\mathbf{x}_{i}-\mathbf{c}_{X}\Vert_{2}$ and $\mathbf{x}_{\beta}=\arg\max_{\mathbf{x}_{i}\in X}\Vert\mathbf{x}_{i}-\mathbf{x}_{\alpha}\Vert_{2}$. Then, a GB $\Omega_{X}$ can be divided into GBs $\Omega_{X_{\alpha}}$ and $\Omega_{X_{\beta}}$, where 
\begin{align}\label{Equation:Sub-Granular-Balls}
    X_{\alpha}=\left\{\mathbf{x}_{i}\in X:\Vert\mathbf{x}_{i}-\mathbf{x}_{\alpha}\Vert_{2}\leq\Vert\mathbf{x}_{i}-\mathbf{x}_{\beta}\Vert_{2}\right\},
    X_{\beta}=X/X_{\alpha}.
\end{align}
Throughout the subsequent discussions, GBs $\Omega_{X_{\alpha}}$ and $\Omega_{X_{\beta}}$ are invariably called sub-GBs of $\Omega_{X}$. Moreover, in \cite{JiaZihang2025TCYB}, a GB $\Omega_{X}$ is considered sufficiently small, if it holds that $\vert X\vert\leq\delta\cdot\sqrt{n}$, where $\delta\in]0,1]$ is a parameter. 

During the pre-division process of GBs, a GB-based binary tree $\mathcal{T}_{U}$ rooted at GB $\Omega_{U}$ can be naturally established. In tree $\mathcal{T}_{U}$, for any GB that is insufficiently small, its sub-GBs, as defined by Eq. (\ref{Equation:Sub-Granular-Balls}), are designated as its child nodes. The decision to further divide or retain a node in $\mathcal{T}_{U}$ as a generated GB is based on its best quality. Moreover, the GBs generated from universe $U$ are determined by calculating the best combination of sub-GBs of GB $\Omega_{U}$. Specifically, the best quality and the best combination of sub-GBs for a GB are defined as below.  

\begin{defn}\label{Definition:Best Quality and Best Combination of Sub-Granular-Balls of Granular-Ball} (See \cite{JiaZihang2025TCYB})
Let $\mathcal{T}_{U}$ be the GB-based binary tree established by GB-POJG, GB $\Omega_{X}$ be a node in tree $\mathcal{T}_{U}$, with sub-GBs $\Omega_{X_{\alpha}}$ and $\Omega_{X_{\beta}}$ determined by Eq. (\ref{Equation:Sub-Granular-Balls}).  
\begin{enumerate}
\item[(1)] If $\Omega_{X}$ is a non-leaf node, its best quality level $\mathcal{BQ}\left(\Omega_{X}\right)$ is defined as $\mathcal{BQ}\left(\Omega_{X}\right)=\max\left(\mathcal{Q}\left(\Omega_{X}\right),\mathcal{BQ}\left(\Omega_{X_{\alpha}}\right)+\mathcal{BQ}\left(\Omega_{X_{\beta}}\right)\right)$. Otherwise, $\mathcal{BQ}\left(\Omega_{X}\right)=\mathcal{Q}\left(\Omega_{X}\right)$.
\item[(2)] If it holds that $\mathcal{BQ}\left(\Omega_{X}\right)=\mathcal{Q}\left(\Omega_{X}\right)$, then the best combination of sub-GBs of $\Omega_{X}$, denoted by $\mathcal{BC}\left(\Omega_{X}\right)$, is given by $\mathcal{BC}\left(\Omega_{X}\right)=\left\{\Omega_{X}\right\}$. Otherwise, $\mathcal{BC}\left(\Omega_{X}\right)=\mathcal{BC}\left(\Omega_{X_{\alpha}}\right)\cup \mathcal{BC}\left(\Omega_{X_{\beta}}\right)$. 
\end{enumerate} 
\end{defn}

Since the absence of instance labels may result in some GBs being positioned at decision boundaries or containing noise, potentially resulting in inaccurate decision boundaries, GB-POJG includes a mechanism to detect abnormal GBs. Let $\Phi=\{\Omega_{X_{1}},\cdots,\Omega_{X_{t}}\}$ be a set of generated GBs. Table \ref{Table:Existing Definition for abnormal GBs} summarizes three exsting criteria for identifying abnormal GBs, where $\mathbf{R}^{\text{ave}}_{\Phi}=\left[R^{\text{ave}}_{X_{1}},\cdots,R^{\text{ave}}_{X_{t}}\right]$, $\mathbf{R}^{\text{max}}_{\Phi}=\left[R^{\text{max}}_{X_{1}},\cdots,R^{\text{max}}_{X_{t}}\right]$, and $\mathbf{N}_{\Phi}=\left[\vert X_{1}\vert,\cdots,\vert X_{t}\vert\right]$. The anomaly detection process  iteratively divides abnormal GBs based on a criterion specified in Table \ref{Table:Existing Definition for abnormal GBs}, and this process continues until all abnormal GBs have been divided.

\begin{table}[!htbp]
  \centering
  \fontsize{6}{6}\selectfont
  \renewcommand\tabcolsep{15pt}
  \begin{threeparttable}
  \caption{Existing Criteria for Detecting Abnormal GBs}
  \label{Table:Existing Definition for abnormal GBs}
    \begin{tabular}{ll}
    \toprule
  Reference & Criterion for Detecting Abnormal GBs\\
    \midrule
    \cite{XieJiang2023TKDE}&$R_{X_{i}}^{\text{max}}>2\cdot\max\left(\text{mean}\left(\mathbf{R}_{\Phi}^{\text{max}}\right),\text{median}\left(\mathbf{R}_{\Phi}^{\text{max}}\right)\right)$\\
    \cite{XieJiang2024ICDE}&$R_{X_{i}}^{\text{max}}>1.5\cdot\max\left(\text{mean}\left(\mathbf{R}_{\Phi}^{\text{max}}\right),\text{median}\left(\mathbf{R}_{\Phi}^{\text{max}}\right)\right)$\\
    \cite{JiaZihang2025TCYB}&$R_{X_{i}}^{\text{ave}}>2\cdot\text{mean}\left(\mathbf{R}_{\Phi}^{\text{ave}}\right)$ and $\vert X_{i}\vert<0.5\cdot\text{mean}\left(\mathbf{N}_{\Phi}\right)$\\
    
    \bottomrule
    \end{tabular}
    \end{threeparttable}
\end{table}

\subsection{Clustering Granular-Balls Based on Density}\label{Section2.3}
In this section, a detailed description of the GBDPC algorithm is provided \cite{ChengDongdong2024TNNLS}. GBDPC clusters the generated GBs through the following four main steps \cite{ChengDongdong2024TNNLS}: (1) calculating the density of each GB; (2) computing the pairwise distances between GBs; (3) selecting GBs with peak density as cluster centers; (4) assigning each non-center GB to the cluster of its nearest neighbor with higher density. Clearly, the key distinction between GBDPC and the original DPC algorithm lies in the choice of computational units: GBDPC operates on GBs rather than individual data instances. In \cite{ChengDongdong2024TNNLS}, the density of a GB $\Omega_{X}$ is defined as
\begin{align}\label{Equation:Granular-Ball-Density-GBDPC}
  \rho\left(\Omega_{X}\right)=\vert X\vert\cdot\left(R_{X}^{\text{ave}}\cdot\left(R_{X}^{\text{max}}\right)^{2}\right)^{-1},
\end{align}
while the Euclidean distance between GB centers is employed to measure inter-GB distance. To better capture the geometric characteristics of GBs, an alternative distance metric is adopted in this paper, as shown below:
\begin{defn}\label{Definition:Distance Between Granular-Balls} (See \cite{XiaShuyin2019INS})
The distance between two GBs $\Omega_{X}$ and $\Omega_{Y}$, denoted as $\mathcal{D}(\Omega_{X},\Omega_{Y})$, is defined as 
\begin{align}\label{Equation:Distance-Between-Granular-Balls}
  \mathcal{D}\left(\Omega_{X}, \Omega_{Y}\right)=\max\left(0,\Vert\mathbf{c}_{X}-\mathbf{c}_{Y}\Vert_{2}-\left(R_{X}^{\text{ave}}+R_{Y}^{\text{ave}}\right)\right).
\end{align}
\end{defn}

\section{Details of the LGBQPC Algorithm}\label{Section3}

This section provides a comprehensive description of the proposed LGBQPC algorithm. Specifically, Section \ref{Section3.1} introduces an enhanced GB generation method tailored for clustering tasks, referred to as GB-POJG+. Section \ref{Section3.2} then presents the GB-based clustering method utilizing the $k$-NN graph. Finally, the complete LGBQPC algorithm along with its time complexity analysis is detailed in Section \ref{Section3.3}.

\subsection{GB-POJG+}\label{Section3.1}

This section presents the details of the proposed GB-POJG+ method, which is developed as an enhanced version of GB-POJG. GB-POJG+ incorporates four key enhancements:  (1) a revised objective function, (2) a novel termination criterion for GB division, (3) a redefinition of abnormal GBs, and (4) an adaptive strategy for granularity level control.

\subsubsection{Objective Function}\label{Section3.1.1} The original GB-POJG method is designed to maximize the overall quality of the generated GBs. However, it does not account for the influence of the number of GBs on the subsequent clustering process. In fact, generating too many GBs can degrade computational efficiency and reduce robustness. To address this issue, a new objective function is proposed for GB-POJG+, formulated as:
\begin{align}\label{Equation:Objective-Function-GB-POJG+}
J^{*}(\Phi,\lambda)
=&\left(\sum\nolimits_{i=1}^{t}\mathcal{Q}(\Omega_{X_{i}})\right)-\lambda t,
\end{align}
where $\Phi=\{\Omega_{X_{1}},\cdots,\Omega_{X_{t}}\}$ denotes the set of generated GBs, and $\lambda\geq0$ corresponds the penalty coefficient that regulates the number of generated GBs. This formulation allows for maximizing the overall quality of GBs while while keeping their quantity within a reasonable range. Notably, when $\lambda=0$, Eq. (\ref{Equation:Objective-Function-GB-POJG+}) degenerates into the original objective function of GB-POJG (i.e., Eq. \eqref{Equation:Objective-Function-GB-POJG}).

\subsubsection{Termination Criterion for GB Division}\label{Section3.1.2} To effectively maximize the revised objective function in Eq. (\ref{Equation:Objective-Function-GB-POJG+}), it is imperative to establish a suitable termination criterion for GB division. Rewriting the objective function yields: 
\begin{align}\label{Equation:Revisit-Objective-Function-GB-POJG+}
  J^{*}(\Phi,\lambda)
  =&\sum\nolimits_{i=1}^{t}\left(\mathcal{Q}(\Omega_{X_{i}})-\lambda\right).
\end{align}
Here, each term $\mathcal{Q}(\Omega_{X_{i}})-\lambda$ can be treated as an individual penalized quality level. Structurally, this form is similar to the original objective in GB-POJG, indicating that similar techniques can be employed for optimization. Consequently, the penalized quality level of a GB can be defined as follows:
\begin{defn}\label{Definition:Penalized-Granular-Ball-Quality}
  Let $\lambda\geq0$ be a penalty coefficient. The penalized quality level of a GB $\Omega_{X}$, denoted by $\mathcal{PQ}(\Omega_{X},\lambda)$, is defined as $\mathcal{PQ}(\Omega_{X},\lambda)=\mathcal{Q}(\Omega_{X})-\lambda$, where $\mathcal{Q}(\Omega_{X})$ is the quality level of $\Omega_{X}$ defined in Definition \ref{Definition:Quality of Granular-Ball}.
\end{defn}

To maximize the objective function Eq. (\ref{Equation:Objective-Function-GB-POJG+}), GB-POJG+ adopts a two-stage, parameter-free strategy similar to GB-POJG in its initial stage. This strategy involves GB pre-division and the construction of a binary tree $\mathcal{T}_{U}$ of GBs, as outlined below:
\begin{description}
  \item [\bf Step 1]: Starting from the initial GB $\Omega_{U}$ derived from the universe $U$, iteratively divide each GBs until it is sufficiently small, i.e., the number of instances belonging to each GB falls below a threshold $\sqrt[3]{n}$. Subsequently, a binary tree $\mathcal{T}_{U}$ is then constructed by designating the sub-GBs of each insufficiently small GB as its child nodes.
  \item [\bf Step 2]: Detect and completely divide abnormal leaf nodes in the tree $\mathcal{T}_{U}$ established in \textbf{Step 1}. 
\end{description}

Specifically, both GB-POJG and GB-POJG+ share the common objective of ensuring that each leaf node in $\mathcal{T}_{U}$ is derived from instances belonging to the same cluster. However, GB-POJG achieves this by enforcing a size threshold (less than $\delta \cdot \sqrt{n}$) for leaf nodes. Although reducing this threshold enables finer division for GBs and better objective attainment, it simultaneously increases the number of GBs generated by GB-POJG \cite{JiaZihang2025TCYB}. Consequently, to balance clustering efficiency and effectiveness, previous work recommends setting the threshold within $[0.4\sqrt{n}, \sqrt{n}]$ \cite{JiaZihang2025TCYB}. In contrast, GB-POJG+ avoids over-generation of GBs by directly optimizing its objective function (i.e., Eq. \eqref{Equation:Objective-Function-GB-POJG+}), allowing it to adopt a lower threshold without sacrificing clustering quality. Notably, it holds that $0.4\sqrt{n}>\sqrt[3]{n}$ for $n\geq245$. This indicates that $\sqrt[3]{n}$ consistently falls below the recommended threshold range for GB-POJG in  medium- to large-scale datasets. Hence, GB-POJG+ employs $\sqrt[3]{n}$ as a parameter-free threshold for the threshold for defining sufficiently small GBs. Furthermore, GB-POJG+ integrates anomaly detection during tree construction to mitigate the adverse of leaf nodes located at decision boundaries or containing noise. Here, the complete division of a GB refers to recursively dividing it until each resulting GB contains only a single instance.  Meanwhile, the binary tree $T_{U}$ should be updated by designating sub-GBs of each GB as its child nodes.

To further identify which GBs should be retained in the final set, each GB's penalized best quality and penalized best combination of sub-GBs are formally defined in accordance with Definition \ref{Definition:Best Quality and Best Combination of Sub-Granular-Balls of Granular-Ball} as follows.

\begin{defn}\label{Definition:Penalized-Best-Quality-and-Combination}
  Let $\mathcal{T}_{U}$ be the binary tree constructed by GB-POJG+, $\lambda\geq0$ be the penalty coefficient, $\Omega_{X}$ be a node in tree $\mathcal{T}_{U}$, $\mathcal{PQ}(\Omega_{X},\lambda)$ be the penalized quality level of $\Omega_{X}$, and $\Omega_{X_{\alpha}}$ and $\Omega_{X_{\beta}}$ be sub-GBs of $\Omega_{X}$ determined by Eq. (\ref{Equation:Sub-Granular-Balls}).
  \begin{enumerate}
    \item [(1)] If $\Omega_{X}$ is a leaf node, its penalized best quality level, denoted by $\mathcal{PBQ}(\Omega_{X},\lambda)$, is defined as $\mathcal{PBQ}(\Omega_{X},\lambda)=\mathcal{PQ}(\Omega_{X},\lambda)$. Otherwise, $\mathcal{PBQ}(\Omega_{X},\lambda)=\max\left(\mathcal{PQ}(\Omega_{X},\lambda), \mathcal{PBQ}(\Omega_{X_{\alpha}},\lambda)+\mathcal{PBQ}(\Omega_{X_{\beta}},\lambda)\right)$.
    \item [(2)] If $\mathcal{PBQ}(\Omega_{X},\lambda)=\mathcal{PQ}(\Omega_{X},\lambda)$, then the penalized best combination of sub-GBs of $\Omega_{X}$, denoted by $\mathcal{PBC}(\Omega_{X},\lambda)$, is $\left\{\Omega_{X}\right\}$. Otherwise, $\mathcal{PBC}(\Omega_{X},\lambda)=\mathcal{PBC}(\Omega_{X_{\alpha}},\lambda)\cup \mathcal{PBC}(\Omega_{X_{\beta}},\lambda)$.
  \end{enumerate}
\end{defn}

In summary, GB-POJG+ begins by generating an initial set of GBs through a two-stage, parameter-free division process and constructs a binary tree $\mathcal{T}{U}$. Following this, for each GB node in $\mathcal{T}_{U}$, if its penalized best quality exceeds its own penalized quality, it is further divided; otherwise, it is retained. Ultimately, the set of generated GBs are determined by the penalized best combination of sub-GBs for the root GB $\Omega_{U}$.

\subsubsection{Definition of Abnomal GBs}\label{Section3.1.3} Following previous studies \cite{XieJiang2023TKDE,XieJiang2024ICDE,JiaZihang2025TCYB}, GB-POJG+ iteratively detects and divides anomalies within the generated GBs, thereby improving decision boundary accuracy in subsequent clustering. Notably, the construction of the GB-based binary tree also involves the identification of abnormal GBs. Consequently, the performance of GB-POJG+ heavily  depends on the rationality of the abnormal GB  definition. 

However, as shown in Table \ref{Table:Existing Definition for abnormal GBs}, existing definitions typically consider only either the maximum or average radius of GBs. Moreover, thresholds for identifying abnormal GBs are often based on fixed multiples of the mean or median, making them sensitive to both outliers and data scale. To address these limitations, a refined definition of an abnormal GB is presented as follows. 
\begin{defn}\label{Definition:Abnormal-Granular-Balls-Sigma}
Let $\Phi=\{\Omega_{X_{1}},\cdots,\Omega_{X_{t}}\}$ be a set of GBs, $\mathbf{R}^{\text{ave}}_{\Phi}=\left[R^{\text{ave}}_{X_{1}},\cdots,R^{\text{ave}}_{X_{t}}\right]$, $\mathbf{R}^{\text{max}}_{\Phi}=\left[R^{\text{max}}_{X_{1}},\cdots,R^{\text{max}}_{X_{t}}\right]$, and $\mathbf{N}_{\Phi}=\left[\vert X_{1}\vert,\cdots,\vert X_{t}\vert\right]$. For each $\Omega_{X_{i}}\in\Phi~(i=1,\cdots,t)$, it is deemed an abnormal GB, if one of the following conditions is met: (1) $R_{X_{i}}^{\text{max}}>\text{mean}(\mathbf{R}^{\text{max}}_{\Phi})+\text{std}(\mathbf{R}^{\text{max}}_{\Phi})$; (2) $R_{X_{i}}^{\text{ave}}>\text{mean}(\mathbf{R}^{\text{ave}}_{\Phi})+\text{std}(\mathbf{R}^{\text{ave}}_{\Phi})$ and $\vert X_{i}\vert<\text{mean}(\mathbf{N}_{\Phi})-\text{std}(\mathbf{N}_{\Phi})$.
\end{defn}

\begin{rem}
Definition \ref{Definition:Abnormal-Granular-Balls-Sigma} introduces a more comprehensive criterion for identifying abnormal GBs. Specifically, Condition (1), inspired by \cite{XieJiang2023TKDE,XieJiang2024ICDE}, identifies abnormality based on the maximum radius of a GB, while Condition (2), motivated by \cite{JiaZihang2025TCYB}, considers both the average radius and the number of instances within a GB. Additionally, by incorporating both the mean and standard deviation, Definition \ref{Definition:Abnormal-Granular-Balls-Sigma} adaptively set thresholds, enhancing robustness in abnormal GB detection. The effectiveness of Definition \ref{Definition:Abnormal-Granular-Balls-Sigma} is further validated through experiments presented in Section \ref{Section:Ablation-Study}.
\end{rem}

\subsubsection{Adaptation Strategy for Granularity Level}\label{Section3.1.4}
GB-POJG+ utilizes the GB quality metric defined in Definition \ref{Definition:Quality of Granular-Ball}. Although this definition offers a comprehensive framework for assessing the quality of GBs, it involves a granularity level parameter that must be carefully tuned to balance coverage and specificity (see Remark \ref{Remark:Granularity-Level}). To eliminate the need for manual tuning, an adaptive granularity level adjustment strategy is introduced, as detailed below.

The fundamental principle of this strategy is to determine an appropriate granularity level that maximizes the division of GBs satisfying two criteria: (1) the GBs are not sufficiently small, and (2) the instances within them cannot be effectively represented by a hyperspherical structure. 

\begin{figure}[!htbp]
  \centering
  \subfigure[]{\includegraphics[width=0.32\linewidth]{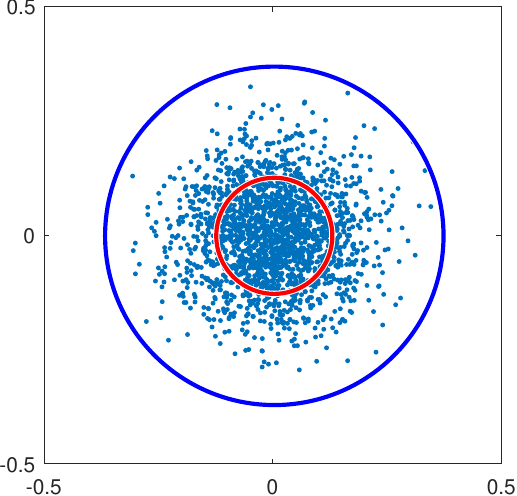}}
  \subfigure[]{\includegraphics[width=0.32\linewidth]{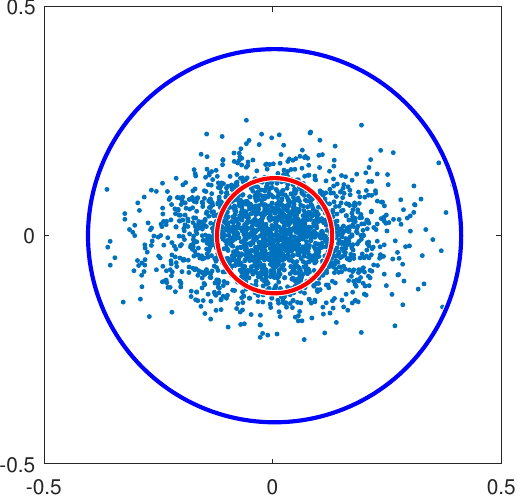}}
  \subfigure[]{\includegraphics[width=0.32\linewidth]{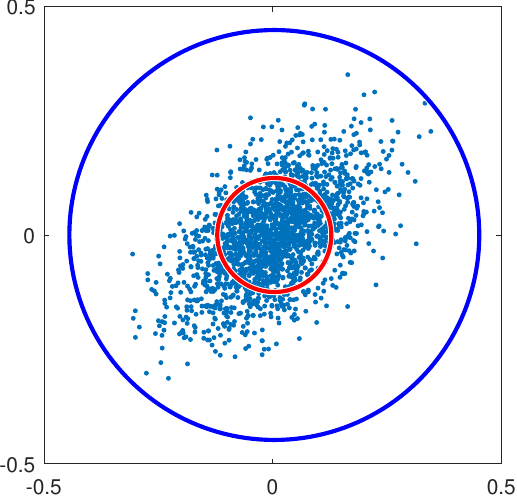}}
  \caption{Three GBs with average (red) and maximum (blue) radii.}
  \label{Figure: GBs derived from instances with different distribution}
\end{figure}

Specifically, in GB-POJG+, most sufficiently small GBs must be retained without further division. Additionally, since the GB computing framework uses hyperspheres to represent instance distributions, any deviation from this assumption may misalign the resulting GBs with the actual data distribution. Consider the synthetic datasets shown in Figs. \ref{Figure: GBs derived from instances with different distribution}(a)-(c). Clearly, the instances shown in Fig. \ref{Figure: GBs derived from instances with different distribution}(a) are well-represented by a hypersphere. In contrast, those in Figs. \ref{Figure: GBs derived from instances with different distribution}(b) and \ref{Figure: GBs derived from instances with different distribution}(c) exhibit significant limitations for such representation due to significant variation in feature dispersion  and strong feature correlation, respectively.  In GB-POJG+, feature dispersion is quantified by the coefficient of variation in feature variances, while feature correlation is measured by the absolute value of the correlation coefficient. According to empirical experience, a coefficient of variation below 0.1 indicates low dispersion \cite{Lande1977SZ}, and a correlation coefficient below 0.3 indicates a relatively weak inter-feature correlation \cite{Ross2014IPSES}. These criteria guide the formal definition below.
\begin{defn}\label{Definition:Granularity-Level-Adjustment-Criterion}
A GB $\Omega_{X}$ is said to satisfy the granularity level adjustment criterion, if (1) $\vert X\vert\geq\sqrt[3]{n}$, and (2) the instances within $\Omega_{X}$ exhibit either a coefficient of variation in feature variances exceeding 0.1 or an absolute correlation coefficient between features greater than 0.3.
\end{defn}

Furthermore, according to Definition \ref{Definition:Penalized-Best-Quality-and-Combination}, any non-leaf node in the tree $\mathcal{T}_{U}$ must be divided, if it satisfies:
\begin{align}
  \mathcal{PQ}(\Omega_{X},\lambda)&<\mathcal{PQ}(\Omega_{X_{\alpha}},\lambda)+\mathcal{PQ}(\Omega_{X_{\beta}},\lambda)\label{Equation:Inequality-GB-Sub-GBs-Granularity-Level}\\
  &\leq\mathcal{PBQ}(\Omega_{X_{\alpha}},\lambda)+\mathcal{PBQ}(\Omega_{X_{\beta}},\lambda).\notag
\end{align}
In conjunction with Definitions \ref{Definition:Quality of Granular-Ball}, \ref{Definition:Penalized-Granular-Ball-Quality}, and \ref{Definition:Penalized-Best-Quality-and-Combination}, and Remark \ref{Remark:Granularity-Level}, Eq. (\ref{Equation:Inequality-GB-Sub-GBs-Granularity-Level}) can be reformulated as:
\begin{align}\label{Equation:Transcendental-Inequality-GB-Sub-GBs-Granularity-Level}
  \frac{\mathcal{Q}_{C}(\Omega_{X})}{\exp(\gamma\cdot R^{\text{ave}}_{X})}
  <\frac{\mathcal{Q}_{C}(\Omega_{X_{\alpha}})}{\exp(\gamma\cdot R^{\text{ave}}_{X_{\alpha}})}+\frac{\mathcal{Q}_{C}(\Omega_{X_{\beta}})}{\exp(\gamma\cdot R^{\text{ave}}_{X_{\beta}})}-\lambda.
\end{align}
Since Eq. (\ref{Equation:Transcendental-Inequality-GB-Sub-GBs-Granularity-Level}) is a transcendental inequality in terms of the granularity level $\gamma$ and typically lacks an analytical solution, this paper adopts the following approximations as in Definition \ref{Definition:Quality of Granular-Ball}:
\begin{align}\label{Equation:Increasing-Decreasing-Functions}
  f_{1}(t)=t~\text{and}~f_{2}(t)=(1+\gamma\cdot t)^{-1}.
\end{align} 
Substituting these into Eq. (\ref{Equation:Inequality-GB-Sub-GBs-Granularity-Level}) yields:
\begin{align}\label{Equation:Cubic-Inequality-GB-Sub-GBs-Granularity-Level}
  \frac{\mathcal{Q}_{C}(\Omega_{X})}{1+\gamma\cdot R^{\text{max}}_{X}}<\frac{\mathcal{Q}_{C}(\Omega_{X_{\alpha}})}{1+\gamma\cdot R^{\text{max}}_{X_{\alpha}}}+\frac{\mathcal{Q}_{C}(\Omega_{X_{\beta}})}{1+\gamma\cdot R^{\text{max}}_{X_{\beta}}}-\lambda,
\end{align}
which is a tractable cubic inequality that permits analytical solutions. Following the global-first cognitive mechanism \cite{WangGuoyin2017GrC}, the adaptation strategy prioritizes GBs encountered earlier in the pre-division stage. Consequently, the full strategy is described below:

\begin{description}
  \item [\bf Step 1]: Initialize the granularity level range as $\overline{\gamma}=[0,+\infty[$.
  \item [\bf Step 2]: Perform a breadth-first search starting from the root node of the GB-based binary tree established by GB-POJG+. For each GB encountered during the search that satisfies the granularity level adjustment criterion in Definition \ref{Definition:Granularity-Level-Adjustment-Criterion}, solve Eq. (\ref{Equation:Cubic-Inequality-GB-Sub-GBs-Granularity-Level}) to determine the solution set $\gamma^{*}$. Whenever $\overline{\gamma}\cap\gamma^{*}\neq\emptyset$, update $\overline{\gamma}$ to $\overline{\gamma}\cap\gamma^{*}$.
  \item [\bf Step 3]: Set the granularity level $\gamma$ as $\gamma=\inf_{\theta \in\overline{\gamma}}\theta+\varepsilon$, where $\varepsilon$ is a sufficiently small positive constant.
\end{description}

\begin{algorithm2e}[!htbp]
  \footnotesize
  % \SetAlgoLined			
  \DontPrintSemicolon		
      \SetKwInOut{Input}{\textbf{Input}}		
      \SetKwInOut{Output}{\textbf{Output}}	
  \Input{Universe $U=\{\mathbf{x}_{1},\cdots,\mathbf{x}_{n}\}$, number of clusters $c$, penalty coefficient $\lambda$, and number of neighbors $k$.} 

  \Output{Set of instance labels $\mathcal{L}=\{\mathcal{L}(\mathbf{x}_{1}),\cdots,\mathcal{L}(\mathbf{x}_{n})\}$.}

  \textbf{Initialize:} a tree $\mathcal{T}_{U}$ rooted at the GB $\Omega_{U}$, granularity level range $\overline{\gamma}\gets[0,+\infty[$, and a sufficiently small positive constant $\varepsilon$.\\

  \tcc{ Initial GB generation using GB-POJG+}
  \While{$\exists\Omega_{X}\in\mathcal{T}_{U}$ such that $\vert X\vert>\sqrt[3]{n}$, and $\Omega_{X}$ is a leaf node}{
    \tcp{perform breadth-first search}
    divide $\Omega_{X}$ into $\Omega_{X_{\alpha}}$ and $\Omega_{X_{\beta}}$ using Eq. (\ref{Equation:Sub-Granular-Balls});\\
    designate $\Omega_{X}$ as the parent node of $\Omega_{X_{\alpha}}$ and $\Omega_{X_{\beta}}$;\\
    \If{$\Omega_{X}$ satisfies the granularity level adjustment criterion in Definition \ref{Definition:Granularity-Level-Adjustment-Criterion}}{
      $\gamma^{*}\gets$ the solution to Eq. (\ref{Equation:Cubic-Inequality-GB-Sub-GBs-Granularity-Level}) for $\Omega_{X}$;\\
      \If{$\overline{\gamma}\cap\gamma^{*}\neq\emptyset$}{
        $\overline{\gamma}\gets\overline{\gamma}\cap\gamma^{*}$;
      }
    }
  }
  $\gamma\gets\inf_{\theta\in\overline{\gamma}}\theta+\varepsilon$;\tcp*{obtain granularity level}
  detect abnormal GBs among all leaf nodes in $\mathcal{T}_{U}$ using Definition \ref{Definition:Abnormal-Granular-Balls-Sigma};\\
  \ForEach{leaf node $\Omega_{X}\in\mathcal{T}_{U}$}{
    \If{$\Omega_{X}$ is abnormal}{
      fully divide $\Omega_{X}$ using Eq. (\ref{Equation:Sub-Granular-Balls}) and update $\mathcal{T}_{U}$;
    }
  }
  \ForEach{$\Omega_{X}\in\mathcal{T}_{U}$}{
    calculate $\mathcal{PBQ}(\Omega_{X})$ and $\mathcal{PBC}(\Omega_{X})$ by Definition \ref{Definition:Penalized-Best-Quality-and-Combination};\\
  }
  $\Phi\gets\mathcal{PBC}(\Omega_{U})$;\\
  detect abnormal GBs in $\Phi$ using Definition \ref{Definition:Abnormal-Granular-Balls-Sigma};\\
  \While{$\exists\Omega_{X}\in\Phi$ is abnormal}{
    $\Phi\gets\left(\Phi/\left\{\Omega_{X}\right\}\right)\cup\left\{\Omega_{X_{\alpha}},\Omega_{X_{\beta}}\right\}$ using Eq. (\ref{Equation:Sub-Granular-Balls});\\
  }
  \tcc{Clustering based on the GB $k$-NN graph}
  construct the GB $k$-NN graph $\mathcal{G}_{k}=(\Phi,\mathcal{E})$ by Definition \ref{Definition:Granular-Balls-k-Nearest-Neighbors-Graph};\\
  calculate the relative quality for each $\Omega_{X}\in\Phi$ using Definition \ref{Definition:Relative-Quality-Granular-Balls};\\
  calculate the geodesic distance between each pair of GBs in $\Phi$ using Definition \ref{Definition:Geodesic-Distance-Granular-Balls};\tcp*{using Dijkstra's algorithm}
  calculate the relative geodesic distance and relative nearest neighbor for each $\Omega_{X}\in\Phi$ using Definition \ref{Definition:Relative-Geodesic-Distance-Nearest-Neighbor};\\
  calculate the decision value for each $\Omega_{X}\in\Phi$ using Definition \ref{Definition:Decision-Value};\\
  sort all GBs in $\Phi$ in descending order of decision values,  resulting in the ordered set $\Phi=\left\{\Omega_{X_{i_{1}}},\cdots,\Omega_{X_{i_{p}}}\right\}$;\\
  \For{$j=1$ to $p$}{
    \eIf{$1\leq j\leq c$}{
      $\mathcal{L}_{GB}\left(\Omega_{X_{i_{j}}}\right)\gets j$; 
    }{
      $\mathcal{L}_{GB}\left(\Omega_{X_{i_{j}}}\right)\gets\mathcal{L}_{GB}\left(\mathcal{N}_{R}\left(\Omega_{X_{i_{j}}}\right)\right)$; 
    }
    \ForEach{$\mathbf{x}_{q}\in X_{i_{j}}$}{
      $\mathcal{L}(\mathbf{x}_{q})\gets\mathcal{L}_{GB}\left(\Omega_{X_{i_{j}}}\right)$;
    }
  }
  \textbf{return:} the set of instance labels $\mathcal{L}=\{\mathcal{L}(\mathbf{x}_{1}),\cdots,\mathcal{L}(\mathbf{x}_{n})\}$;
  \caption{LGBQPC}
  \label{Algorithm:LGBQPC}
\end{algorithm2e}

\subsection{Clustering Granular-Balls Based on Local Quality Peaks}\label{Section3.2}

In GBDPC, the GB density is calculated using Eq. (\ref{Equation:Granular-Ball-Density-GBDPC}), which exhibits two critical limitations: (1) GBs composed of only a few closely clustered noise instances may be assigned excessively high density values, and (2) GB density is estimated from a global perspective, disregarding local neighborhood information. These shortcomings can result in erroneous clustering results, particularly on datasets with non-uniform density distributions. Moreover, GBDPC  measures the distance between GBs using the Euclidean distance between their centers, which fails to capture the geometric structure of GBs and identify complex manifold structures. To overcome these limitations and further enhance the performance of GBDPC, this section redefines both the density estimation and distance metrics for GBs. 

To begin with, a new method for density estimation is proposed. As shown in Definition \ref{Definition:Quality of Granular-Ball} and Eq. (\ref{Equation:Granular-Ball-Density-GBDPC}), both GB quality and density are positively correlated with the number of instances within the GB and negatively correlated with its radius. Moreover, the GB quality, as specified by Definition \ref{Definition:Quality of Granular-Ball} and Eq. (\ref{Equation:Increasing-Decreasing-Functions}), is bounded within the interval $]0,\mathcal{Q}_{C}(\Omega_{X})]$. These findings imply that utilizing this quality metric for GB density estimation can effectively avoid overestimating density. Additionally, inspired by \cite{HouJian2020TII}, a cluster center is required to exhibit only locally higher density compared to its neighbors rather than globally. Accordingly, to handle non-uniform density distributions more effectively, a relative quality metric is introduced for GB density estimation, incorporating both the GB quality metric and neighborhood information. This is achieved by constructing a GB $k$-NN graph for the generated GBs, as defined below.
\begin{defn}\label{Definition:Granular-Balls-k-Nearest-Neighbors-Graph}
  A GB $k$-NN graph is defined as a 2-tuple $\mathcal{G}_{k}=(\Phi,\mathcal{E})$, where $\Phi=\{\Omega_{X_{1}},\cdots,\Omega_{X_{t}}\}$ is the set of GBs, and an edge $(\Omega_{X_{i}},\Omega_{X_{j}})\in\mathcal{E}$ exists if and only if $\Omega_{X_{j}}$ is among the $k$ nearest neighbors of $\Omega_{X_{i}}$ according to the distance metric defined in Eq. (\ref{Equation:Distance-Between-Granular-Balls}).
\end{defn}
Based on Definition \ref{Definition:Granular-Balls-k-Nearest-Neighbors-Graph}, the relative quality metric for a GB is defined as follows.
\begin{defn}\label{Definition:Relative-Quality-Granular-Balls}
  Let $\mathcal{G}_{k}=(\Phi,\mathcal{E})$ be a GB $k$-NN graph. For any $\Omega_{X_{i}}\in\Phi$, its relative quality $\mathcal{RQ}(\Omega_{X_{i}})$ is defined as
  \begin{align}\label{Equation:Relative-Quality-Granular-Balls}
    \mathcal{RQ}(\Omega_{X_{i}})=\frac{\mathcal{Q}(\Omega_{X_{i}})}{\frac{1}{k}\sum\nolimits_{\left(\Omega_{X_{i}},\Omega_{X_{j}}\right)\in\mathcal{E}}\mathcal{Q}(\Omega_{X_{j}})}.
  \end{align}
\end{defn}

Next, a new distance metric for GBs is introduced to further enhance GBDPC. According to \cite{DuMingjing2018IJMLC}, the geodesic distance between instances demonstrates superiority in identifying complex manifold structures and improving DPC performance. Building on this idea, Eq. (\ref{Equation:Distance-Between-Granular-Balls}), which is capable of capturing the geometric structure of GBs, is extended to define the geodesic distance between GBs over the GB $k$-NN graph.
\begin{defn}\label{Definition:Geodesic-Distance-Granular-Balls}
  Let $\mathcal{G}_{k}=(\Phi,\mathcal{E})$ be a GB $k$-NN graph. For any pair $\Omega_{X_{i}},\Omega_{X_{j}}\in\Phi$, the geodesic distance $\mathcal{D}_{G}(\Omega_{X_{i}},\Omega_{X_{j}})$ is the length of the shortest path between them on $\mathcal{G}_{k}$.
\end{defn}

With newly defined density and distance metrics, the criteria for identifying cluster centers and assigning non-center GBs are accordingly refined. To support this refinement, two key concepts are introduced: the relative geodesic distance, which assesses the minimum distance from a GB to any GB with higher relative quality, and the relative nearest neighbor, which identifies the nearest GB with higher relative quality.
\begin{defn}\label{Definition:Relative-Geodesic-Distance-Nearest-Neighbor}
  Let $\mathcal{G}_{k}=(\Phi,\mathcal{E})$ and $\Omega_{X_{i}}\in\Phi$. Define the set of GBs with higher relative quality than $\Omega_{X_{i}}$ as $\mathcal{Y}(\Omega_{X_{i}})=\{\Omega_{X_{j}}\in\Phi:\mathcal{RQ}(\Omega_{X_{j}})>\mathcal{RQ}(\Omega_{X_{i}})\}$. The relative geodesic distance of $\Omega_{X_{i}}$, denoted by $\mathcal{D}_{RG}(\Omega_{X_{i}})$, is defined as 
  \begin{align}\label{Equation:Relative-Geodesic-Distance-Granular-Balls}
    \mathcal{D}_{RG}(\Omega_{X_{i}})=\begin{cases}
      \max\limits_{\Omega_{X_{j}}\in\Phi}\mathcal{D}_{G}(\Omega_{X_{i}},\Omega_{X_{j}}), ~\text{if}~\mathcal{Y}(\Omega_{X_{i}})=\emptyset,\\
      \min\limits_{\Omega_{X_{j}}\in\mathcal{Y}(\Omega_{X_{i}})}\mathcal{D}_{G}(\Omega_{X_{i}},\Omega_{X_{j}}),~\text{otherwise}.\\
    \end{cases}
  \end{align}
  Furthermore, the relative nearest neighbor of $\Omega_{X_{i}}$, denoted by $\mathcal{N}_{R}(\Omega_{X_{i}})$, is defined as 
  \begin{align}\label{Equation:Relative-Nearest-Neighbor-Granular-Balls}
    \mathcal{N}_{R}(\Omega_{X_{i}})=\begin{cases}
      \Omega_{X_{i}}, ~\text{if}~\mathcal{Y}(\Omega_{X_{i}})=\emptyset,\\
      \arg\min\limits_{\Omega_{X_{j}}\in\mathcal{Y}(\Omega_{X_{i}})}\mathcal{D}_{G}(\Omega_{X_{i}},\Omega_{X_{j}}),~\text{otherwise}.\\
    \end{cases}
  \end{align}
\end{defn}

Following the GBDPC framework \cite{ChengDongdong2024TNNLS} and based on Definition \ref{Definition:Relative-Geodesic-Distance-Nearest-Neighbor}, the following refinements are implemented for cluster center identification and non-center GB label assignment:
\begin{itemize}
  \item  A GB is designated as a cluster center (or local quality peak), if it possesses both high relative quality and a large relative geodesic distance.
  \item  A non-center GB inherits the cluster label of its relative nearest neighbor.
  \end{itemize}

 To automate the selection of cluster centers and eliminate manual intervention, a decision value similar to that proposed in \cite{Rodriguez2014Science} is introduced. GBs are ranked in descending order of their decision values, and the top-ranking GBs are designated as cluster centers.
  \begin{defn}\label{Definition:Decision-Value}
    Let $\mathcal{G}_{k}=(\Phi,\mathcal{E})$ be a GB $k$-NN graph. The decision value of GB $\Omega_{X_{i}}\in\Phi$ is defined as 
    \begin{align}
      \mathcal{DV}(\Omega_{X_{i}})=\mathcal{RQ}(\Omega_{X_{i}})\cdot\mathcal{D}_{RG}(\Omega_{X_{i}}),
    \end{align}
    where $\mathcal{RQ}(\Omega_{X_{i}})$ and $\mathcal{D}_{RG}(\Omega_{X_{i}})$ are determined by Eqs. (\ref{Equation:Relative-Quality-Granular-Balls}) and (\ref{Equation:Relative-Geodesic-Distance-Granular-Balls}), respectively.
  \end{defn}

\subsection{The LGBQPC Algorithm and Its Time Complexity Analysis}\label{Section3.3}
\begin{figure*}
  \includegraphics[width=\linewidth]{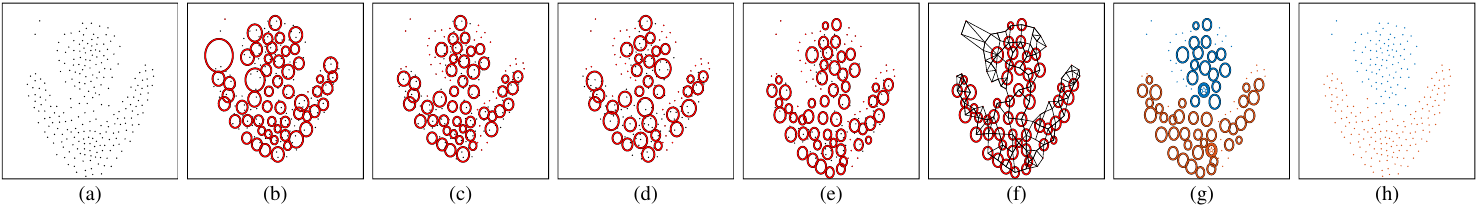}
  \caption{The LGBQPC clustering process illustrated on the Flame dataset \cite{WangYizhang2024ESWA}. (a) The original Flame dataset. (b)-(c) GBs serving as leaf nodes in the binary tree: (b) each leaf node is sufficiently small, and (c) each abnormal leaf node has been completely divided. (d) The best combination of sub-GBs of derived from the Flame dataset. (e) GBs generated after applying anomaly detection to those in (d). (f) The GB $k$-NN graph ($k=3$). (g) Clustering results at the GB level. (h) Final clustering results at the instance level.}
  \label{Figure:Clustering-Process-LGBQPC}
\end{figure*}

The proposed LGBQPC algorithm is outlined in Algorithm \ref{Algorithm:LGBQPC}. It consists of two key components: (1) the generation of GBs using GB-POJG+ and (2) the clustering of the resulting GBs based on the GB $k$-NN graph. An illustrative example of the clustering process is provided in Fig. \ref{Figure:Clustering-Process-LGBQPC}.

The time complexity of Algorithm~\ref{Algorithm:LGBQPC} is analyzed as follows. Let $k$ be the number of neighbors in the GB $k$-NN graph, $t$ be the number of leaf nodes in the tree $\mathcal{T}_{U}$ obtained after the execution of line 8, and $p$ denote the number of GBs generated after executing line 19. If $\mathcal{T}_{U}$ is a full binary tree, the time complexity of lines 2-8 is $O(n\log(t))$, representing the best-case scenario. In contrast, the worst-case time complexity occurs when each left (or right) child node contains only a single instance, leading to a complexity of approximately $O(n^{2})$. According to \cite{JiaZihang2025TCYB}, the operations in lines 9-19, which involve abnormal GB detection and penalized quality evaluation,  incur a time complexity of approximately $O(t)$. Line 20 constructs the GB $k$-NN graph over $p$ GBs, requiring a complexity of $O(p^{2}+kp^{2})$. Line 21 calculates the relative quality for each GB based on its $k$ nearest neighbors, which takes $O(pk)$ time. Line 22 employs Dijkstra's algorithm to calculate the shortest path between all pairs of GBs, resulting in a complexity of $O(kp^2 \log p)$. Line 23, which ranks the GBs by their decision values, has a complexity of $O(p\log p + p^{2})$. Finally, lines 24-32 perform the final labeling based on GBs and instances, with an overall complexity of $O(p+p\log p+n)$. In summary, the overall time complexity of Algorithm \ref{Algorithm:LGBQPC} is approximately $O(n\log t+kp^{2}\log p)$ in the best case. In the worst case, the complexity increases to approximately $O(n^{2}+kp^{2}\log p)$.

\section{Numerical Experiments}\label{Section4}
This section presents extensive numerical experiments to validate the performance of LGBQPC.
\subsection{Experimental Setups}\label{Section4.1}

\subsubsection{Experimental Environment} All experiments are conducted on a workstation running MATLAB 2024b, with system specifications comprising an Intel Core i9-13900HX CPU, 32.0 GB of RAM, and Windows 11 operating system.

\subsubsection{Datasets} 
This study employs a total of 40 datasets, comprising 27 synthetic datasets and 13 publicly available datasets, as detailed in Table \ref{Table:Information of Datasets}. The synthetic datasets are adopted to evaluate the algorithms' ability to cluster data with complex and arbitrary shapes, while the publicly available datasets are used to evaluate their practical applicability. Prior to experimentation, all features are standardized to have zero mean and unit variance to ensure uniformity in scale across all features.
 
\begin{table}[!htbp]
  \centering
  \fontsize{6}{6}\selectfont
  \renewcommand\tabcolsep{3pt}
  \renewcommand\arraystretch{1}
  \begin{threeparttable}
  \caption{Information of Datasets}
  \label{Table:Information of Datasets}
    \begin{tabular}{lllllll}
    \toprule
  ID & Dataset & Source & \#Instances & \#Features & \#Clusters & Type \\
    \midrule
    D1 & Flame & \cite{WangYizhang2024ESWA} & 240 & 2 & 2 & S\\
    D2 & Zelink3 & \cite{WangYizhang2024ESWA} & 266 & 2 & 3 & S\\
    D3 & Zelink1 & \cite{WangYizhang2024ESWA} & 299 & 2 & 3 & S\\
    D4 & Spiral & \cite{WangYizhang2024ESWA} & 312 & 2 & 3 & S\\
    D5 & Jain & \cite{WangYizhang2024ESWA} & 373 & 2 & 2 & S\\
    D6 & 2circles & \cite{WangYizhang2024ESWA} & 600 & 2 & 2 & S\\
    D7 & 2d-3c-no123 & \cite{WangYizhang2024ESWA} & 715 & 2 & 3 & S\\
    D8 & Atom & \cite{WangYizhang2024ESWA} & 800 & 3 & 2 & S\\
    D9 & 2d-4c-no4 & \cite{WangYizhang2024ESWA} & 863 & 2 & 4 & S\\
    D10 & Donut3 & \cite{WangYizhang2024ESWA} & 999 & 2 & 3 & S\\
    D11 & Chainlink & \cite{WangYizhang2024ESWA} & 1000 & 3 & 2 & S\\
    D12 & Smile2 & \cite{WangYizhang2024ESWA} & 1000 & 2 & 4 & S\\
    D13 & Pearl & \cite{WangYizhang2024ESWA} & 1000 & 2 & 2 & S\\
    D14 & Halfkernel & \cite{WangYizhang2024ESWA} & 1000 & 2 & 2 & S\\
    D15 & Donutcurves & \cite{WangYizhang2024ESWA} & 1000 & 2 & 4 & S\\
    D16 & Dartboard2 & \cite{WangYizhang2024ESWA} & 1000 & 2 & 4 & S\\
    D17 & Dartboard1 & \cite{WangYizhang2024ESWA} & 1000 & 2 & 4 & S\\
    D18 & Curves2 & \cite{WangYizhang2024ESWA} & 1000 & 2 & 2 & S\\
    D19 & Crossline & \cite{WangYizhang2024ESWA} & 1000 & 2 & 2 & S\\
    D20 & Crescentfullmoon & \cite{WangYizhang2024ESWA} & 1000 & 2 & 2 & S\\
    D21 & 2spiral & \cite{WangYizhang2024ESWA} & 1000 & 2 & 2 & S\\
    D22 & Clusterincluster & \cite{WangYizhang2024ESWA} & 1012 & 2 & 2 & S\\
    D23 & Complex9 & \cite{WangYizhang2024ESWA} & 3031 & 2 & 9 & S\\
    D24 & Banana & \cite{WangYizhang2024ESWA} & 4811 & 2 & 2 & S\\
    D25 & Algerian Forest Fires (Bejaia) & \cite{Bache2017UCIDatasets} & 122 & 13 & 2 & PA\\
    D26 & Algerian Forest Fires (Sidi-Bel Abbes) & \cite{Bache2017UCIDatasets} & 122 & 13 & 2 & PA\\
    D27 & Iris & \cite{Bache2017UCIDatasets} & 150 & 4 & 3 & PA\\
    D28 & Planning Relax & \cite{Bache2017UCIDatasets} & 182 & 12 & 2 & PA\\
    D29 & Wpbc & \cite{Bache2017UCIDatasets} & 194 & 32 & 2 & PA\\
    D30 & Glass Identification & \cite{Bache2017UCIDatasets} & 214 & 9 & 6 & PA\\
    D31 & Speaker Accent Recognition & \cite{Bache2017UCIDatasets} & 329 & 12 & 6 & PA\\
    D32 & Wholesale Customers & \cite{Bache2017UCIDatasets} & 440 & 7 & 3 & PA\\
    D33 & Blood Transfusion Service Center & \cite{Bache2017UCIDatasets} & 748 & 4 & 2 & PA\\
    D34 & Iranian Churn & \cite{Bache2017UCIDatasets} & 3150 & 13 & 2 & PA\\
    D35 & Abalone & \cite{Bache2017UCIDatasets} & 4177 & 8 & 3 & PA\\
    D36 & Electrical Grid & \cite{Bache2017UCIDatasets} & 10000 & 13 & 2 & PA\\
    D37 & Synthetic Circle & \cite{Bache2017UCIDatasets} & 10000 & 2 & 100 & S\\
    D38 & MAGIC Gamma Telescope & \cite{Bache2017UCIDatasets} & 19020 & 10 & 2 & PA\\
    D39 & TB-60K & \cite{HuangDong2020TKDE} & 60000 & 2 & 2 & S\\
    D40 & CC-60K & \cite{HuangDong2020TKDE} & 60000 & 2 & 3 & S\\
    \bottomrule
    \end{tabular}
    Note: ``S'' and ``PA'' represent synthetic and publicly available, respectively.
    \end{threeparttable}
\end{table}

\subsubsection{Baseline Algorithms and Parameter Settings} To demonstrate the superiority of LGBQPC, it is compared against eleven well-established clustering algorithms: k-means++ \cite{Arthur2007ACM-SIAM}, SC \cite{VonLuxburg2007StatisticsComputing}, DBSCAN \cite{Ester1996KDD}, DPC \cite{Rodriguez2014Science}, DPC-DLP \cite{Seyedi2019ESWA}, FHC-LDP \cite{GuanJunyi2021Neurocomputing}, USPEC \cite{HuangDong2020TKDE}, USENC \cite{HuangDong2020TKDE}, GBDPC \cite{ChengDongdong2024TNNLS}, GBSC \cite{XieJiang2023TKDE}, and GBCT \cite{XiaShuyin2024TNNLS}. 
The parameters for each algorithm are configured as follows:
\begin{itemize}
    \item For SC and GBSC, the affinity matrix is constructed using a $k$-NN graph, where $k=\left\lfloor \log_{2}n\right\rfloor+1$ \cite{VonLuxburg2007StatisticsComputing}.
    \item For DBSCAN, the neighborhood radius is searched within $d_{\max} \cdot \{0, 0.01, \cdots, 0.3\}$, with the minimum number of neighbors required for core points set to $2m$ \cite{Sander1998DMKD}.
   \item For DPC, the truncation distance is selected within $d_{\max} \cdot \{0.02, 0.04, \cdots, 0.5\}$.
   \item For DPC-DLP, the local density parameter is varied within $\{0.1, 0.2, \cdots, 0.5\}$.
  \item For FHC-LDP and LGBQPC, the number of nearest neighbors is selected within $\{1, 2, \cdots, 20\}$ for datasets D1–D38, and within $\{1, 2, \cdots, 30\}$ for datasets D39–D40.
  \item For LGBQPC, the penalty coefficient is searched within $\sqrt[3]{n} \cdot \{0, 0.01, \cdots, 0.3\}$ for datasets D1–D38, and $\sqrt[3]{n} \cdot \{0, 0.005, \cdots, 0.2\}$ for datasets D39–D40.
\end{itemize}

\subsubsection{Evaluation Metrics} To evaluate the performance and efficiency of these clustering algorithm, three widely used metrics, including normalized mutual Information (NMI) \cite{Romano2014ICML}, adjusted rand index (ARI) \cite{Romano2014ICML} and computing overhead (CO) are adopted. Specifically, NMI and ARI fall within the ranges of [0,1] and [-1,1], respectively, whereas CO is measured in seconds. Furthermore, higher values of NMI and ARI signifies better clustering performance, whereas lower values of CO indicates higher clustering efficiency.

\subsubsection{Statistical Tests} To statistically evaluate the performance differences among these algorithms, the Friedman test \cite{Friedman1940ACO} and Nemenyi test \cite{Dunn1961MultipleCA} are conducted. The Friedman test begins with computing the statistic $\tau_{F}$, which follows an $F$-distribution with degrees of freedom $(N_{A}-1)$ and $(N_{A}-1)(N_{D}-1)$:
\begin{align*}
  \tau_{F} =& \frac{(N_{D}-1)\tau_{\mathcal{X}^{2}}}{N_{D}(N_{A}-1)-\tau_{\mathcal{X}^{2}}},\\
  \tau_{\mathcal{X}^{2}}=&\frac{12 N_{D}}{N_{A}(N_{A}+1)}\left(\sum_{i=1}^{N_{A}}r^{2}_{i}-\frac{N_{A}(N_{A}+1)^{2}}{4}\right).
\end{align*}
Here, $N_{A}$, $N_{D}$ and $r_{i}$ denote the number of algorithms being compared, the number of used datasets, and the average rank of $i$-th algorithm across all datasets. Furthermore, given a significance level $\eta$, if $\tau_{F}$ exceeds the critical value $F_{\eta}(N_{A}-1,(N_{A}-1)(N_{D}-1))$, it indicates statistically significant differences among the compared algorithms. Under such circumstances, pairwise differences between the algorithms are subsequently identified using the Nemenyi test. A statistically significant difference is confirmed if the absolute difference between two average ranks exceeds the critical difference $CD_{\eta}=q_{\eta}\sqrt{N_{A}(N_{A}+1)/(6N_{D})}$, where $q_{\eta}$ is the critical value of Tukey's distribution.

\subsection{Experimental Results and Analysis}

\begin{table*}[!htbp]
  \centering
  \fontsize{6}{6}\selectfont
  \renewcommand\tabcolsep{3.8pt}
  \begin{threeparttable}
  \caption{Experimental Results on Datasets D1-D38}
  \label{Table:Experimental Results on Datasets D1-D38}
  \begin{tabular}{llllllllllllll}
    \toprule
    Criteria& Datasets & $k$-means++ & SC     & DBSCAN & DPC    & DPC-DLP & FHC-LDP & USPEC  & USENC  & GBDPC & GBSC   & GBCT   & LGBQPC (Ours) \\
    \midrule
\multirow{39}{*}{NMI ($\uparrow$)} & D1& 39.26 (9.0)& 33.70 (10.0)& 83.13 (5.0)& 96.34 (3.0)& 79.35 (6.0)& \textbf{100.00 (1.5)}& 2.42 (12.0)& 76.42 (7.0)& 26.00 (11.0)& 44.72 (8.0)& 93.59 (4.0)& \textbf{100.00 (1.5)}\\
& D2& 58.17 (12.0)& \textbf{100.00 (3.5)}& \textbf{100.00 (3.5)}& \textbf{100.00 (3.5)}& 89.57 (7.0)& \textbf{100.00 (3.5)}& \textbf{100.00 (3.5)}& 78.19 (9.0)& 65.20 (11.0)& 76.05 (10.0)& 80.71 (8.0)& \textbf{100.00 (3.5)}\\
& D3& 16.24 (12.0)& \textbf{100.00 (2.5)}& \textbf{100.00 (2.5)}& 43.32 (9.0)& 69.78 (7.0)& \textbf{100.00 (2.5)}& 84.06 (5.0)& 66.45 (8.0)& 17.96 (11.0)& 76.54 (6.0)& 22.97 (10.0)& \textbf{100.00 (2.5)}\\
& D4& 0.07 (12.0)& 56.23 (7.0)& \textbf{100.00 (3.0)}& 86.05 (6.0)& 27.46 (8.0)& \textbf{100.00 (3.0)}& \textbf{100.00 (3.0)}& \textbf{100.00 (3.0)}& 3.92 (10.0)& 0.13 (11.0)& 8.86 (9.0)& \textbf{100.00 (3.0)}\\
& D5& 50.29 (8.0)& 53.65 (7.0)& 84.35 (5.0)& \textbf{100.00 (2.5)}& 33.33 (12.0)& \textbf{100.00 (2.5)}& 83.39 (6.0)& 44.93 (9.0)& 43.67 (10.0)& 37.94 (11.0)& \textbf{100.00 (2.5)}& \textbf{100.00 (2.5)}\\
& D6& 0.00 (12.0)& \textbf{100.00 (3.5)}& \textbf{100.00 (3.5)}& 25.50 (10.0)& 35.10 (9.0)& \textbf{100.00 (3.5)}& \textbf{100.00 (3.5)}& 95.44 (7.0)& 2.13 (11.0)& \textbf{100.00 (3.5)}& 37.34 (8.0)& \textbf{100.00 (3.5)}\\
& D7& 69.90 (9.0)& 73.90 (8.0)& 77.09 (6.0)& 88.15 (3.0)& 63.81 (12.0)& 91.01 (2.0)& 80.21 (5.0)& 68.05 (10.0)& 63.83 (11.0)& 75.42 (7.0)& 87.17 (4.0)& \textbf{98.23 (1.0)}\\
& D8& 26.81 (8.0)& \textbf{100.00 (3.0)}& \textbf{100.00 (3.0)}& 43.18 (7.0)& 17.87 (9.0)& \textbf{100.00 (3.0)}& \textbf{100.00 (3.0)}& 96.17 (6.0)& 13.40 (10.0)& 0.00 (12.0)& 0.37 (11.0)& \textbf{100.00 (3.0)}\\
& D9& 82.11 (8.0)& 98.74 (4.0)& 85.95 (7.0)& 99.30 (2.5)& 68.41 (10.0)& 99.30 (2.5)& 61.59 (12.0)& 90.96 (6.0)& 62.71 (11.0)& 98.24 (5.0)& 78.60 (9.0)& \textbf{100.00 (1.0)}\\
& D10& 57.30 (11.0)& \textbf{100.00 (3.0)}& \textbf{100.00 (3.0)}& 70.35 (10.0)& 71.98 (9.0)& \textbf{100.00 (3.0)}& 97.00 (6.0)& 96.10 (7.0)& 52.50 (12.0)& 92.06 (8.0)& \textbf{100.00 (3.0)}& \textbf{100.00 (3.0)}\\
& D11& 19.75 (12.0)& \textbf{100.00 (4.0)}& \textbf{100.00 (4.0)}& 33.22 (10.0)& \textbf{100.00 (4.0)}& \textbf{100.00 (4.0)}& 70.25 (8.0)& 50.93 (9.0)& 24.38 (11.0)& \textbf{100.00 (4.0)}& \textbf{100.00 (4.0)}& \textbf{100.00 (4.0)}\\
& D12& 51.98 (12.0)& \textbf{100.00 (3.5)}& \textbf{100.00 (3.5)}& 76.66 (9.0)& 81.97 (8.0)& \textbf{100.00 (3.5)}& \textbf{100.00 (3.5)}& 95.43 (7.0)& 58.54 (11.0)& \textbf{100.00 (3.5)}& 60.72 (10.0)& \textbf{100.00 (3.5)}\\
& D13& 36.36 (11.0)& 71.86 (7.0)& \textbf{100.00 (2.0)}& 98.12 (5.0)& 41.64 (10.0)& \textbf{100.00 (2.0)}& 85.04 (6.0)& 56.56 (9.0)& 24.34 (12.0)& 63.15 (8.0)& 98.96 (4.0)& \textbf{100.00 (2.0)}\\
& D14& 7.76 (12.0)& \textbf{100.00 (3.5)}& \textbf{100.00 (3.5)}& 47.08 (9.0)& \textbf{100.00 (3.5)}& \textbf{100.00 (3.5)}& 56.44 (7.0)& 33.76 (10.0)& 18.21 (11.0)& \textbf{100.00 (3.5)}& 52.21 (8.0)& \textbf{100.00 (3.5)}\\
& D15& 62.57 (12.0)& \textbf{100.00 (3.0)}& \textbf{100.00 (3.0)}& 84.88 (8.0)& 73.48 (10.0)& \textbf{100.00 (3.0)}& \textbf{100.00 (3.0)}& 97.35 (6.0)& 64.63 (11.0)& 95.07 (7.0)& 80.01 (9.0)& \textbf{100.00 (3.0)}\\
& D16& 52.74 (8.0)& 75.00 (7.0)& \textbf{100.00 (3.0)}& 85.74 (6.0)& 35.58 (10.0)& \textbf{100.00 (3.0)}& \textbf{100.00 (3.0)}& \textbf{100.00 (3.0)}& 15.92 (11.0)& 0.00 (12.0)& 42.06 (9.0)& \textbf{100.00 (3.0)}\\
& D17& 0.00 (12.0)& \textbf{100.00 (3.5)}& \textbf{100.00 (3.5)}& 83.76 (7.0)& 40.51 (10.0)& \textbf{100.00 (3.5)}& \textbf{100.00 (3.5)}& \textbf{100.00 (3.5)}& 3.18 (11.0)& 40.58 (9.0)& 52.80 (8.0)& \textbf{100.00 (3.5)}\\
& D18& 0.00 (12.0)& \textbf{100.00 (5.5)}& \textbf{100.00 (5.5)}& \textbf{100.00 (5.5)}& \textbf{100.00 (5.5)}& \textbf{100.00 (5.5)}& \textbf{100.00 (5.5)}& \textbf{100.00 (5.5)}& 18.50 (11.0)& \textbf{100.00 (5.5)}& \textbf{100.00 (5.5)}& \textbf{100.00 (5.5)}\\
& D19& 0.88 (12.0)& 23.60 (6.0)& 20.22 (7.0)& 30.30 (2.0)& 15.87 (8.0)& 30.03 (3.0)& 5.42 (10.0)& 2.66 (11.0)& 12.42 (9.0)& 26.53 (4.0)& 25.47 (5.0)& \textbf{79.61 (1.0)}\\
& D20& 35.71 (11.0)& \textbf{100.00 (4.5)}& \textbf{100.00 (4.5)}& \textbf{100.00 (4.5)}& \textbf{100.00 (4.5)}& \textbf{100.00 (4.5)}& 67.47 (9.0)& 45.42 (10.0)& 18.96 (12.0)& \textbf{100.00 (4.5)}& \textbf{100.00 (4.5)}& \textbf{100.00 (4.5)}\\
& D21& 2.70 (12.0)& \textbf{100.00 (4.0)}& \textbf{100.00 (4.0)}& \textbf{100.00 (4.0)}& 10.81 (10.0)& \textbf{100.00 (4.0)}& \textbf{100.00 (4.0)}& \textbf{100.00 (4.0)}& 4.11 (11.0)& 30.07 (8.0)& 13.17 (9.0)& \textbf{100.00 (4.0)}\\
& D22& 0.11 (11.0)& \textbf{100.00 (3.5)}& \textbf{100.00 (3.5)}& 49.36 (9.0)& \textbf{100.00 (3.5)}& \textbf{100.00 (3.5)}& 77.79 (8.0)& 93.87 (7.0)& 11.20 (10.0)& \textbf{100.00 (3.5)}& 0.05 (12.0)& \textbf{100.00 (3.5)}\\
& D23& 62.58 (11.0)& 92.92 (3.0)& 89.06 (5.0)& 86.02 (7.0)& 63.26 (10.0)& \textbf{100.00 (1.5)}& 92.60 (4.0)& 88.72 (6.0)& 61.94 (12.0)& 79.68 (8.0)& 70.57 (9.0)& \textbf{100.00 (1.5)}\\
& D24& 32.48 (11.0)& \textbf{100.00 (4.0)}& \textbf{100.00 (4.0)}& 55.70 (9.0)& \textbf{100.00 (4.0)}& \textbf{100.00 (4.0)}& \textbf{100.00 (4.0)}& 99.76 (8.0)& 27.88 (12.0)& \textbf{100.00 (4.0)}& 55.52 (10.0)& \textbf{100.00 (4.0)}\\
& D25& \textbf{33.76 (1.0)}& 9.68 (8.0)& 4.09 (11.0)& 27.76 (3.0)& 17.75 (4.0)& 14.44 (5.0)& 8.27 (10.0)& 8.78 (9.0)& 11.88 (6.0)& 11.87 (7.0)& 0.00 (12.0)& 32.79 (2.0)\\
& D26& 32.03 (2.0)& 3.57 (10.0)& 26.65 (3.0)& 14.25 (5.0)& 16.27 (4.0)& 6.86 (8.0)& 2.47 (11.0)& 12.29 (7.0)& 4.65 (9.0)& 12.60 (6.0)& 0.00 (12.0)& \textbf{34.91 (1.0)}\\
& D27& 64.00 (9.0)& 65.07 (7.0)& 73.37 (5.0)& 77.44 (2.0)& 64.91 (8.0)& \textbf{88.51 (1.0)}& 66.09 (6.0)& 58.76 (10.0)& 54.96 (11.0)& 75.93 (3.0)& 0.00 (12.0)& 74.90 (4.0)\\
& D28& 0.12 (10.0)& \textbf{4.23 (2.0)}& 1.08 (6.0)& \textbf{4.23 (2.0)}& 0.00 (11.5)& 2.13 (4.0)& 0.27 (8.0)& 0.15 (9.0)& 0.36 (7.0)& 1.50 (5.0)& 0.00 (11.5)& \textbf{4.23 (2.0)}\\
& D29& 1.69 (5.0)& 0.99 (9.0)& 1.88 (4.0)& \textbf{8.10 (1.0)}& 0.67 (10.0)& 6.34 (2.0)& 0.48 (11.0)& 1.64 (6.0)& 1.12 (8.0)& 1.14 (7.0)& 0.00 (12.0)& 2.57 (3.0)\\
& D30& 31.01 (6.0)& 30.37 (8.0)& 35.36 (3.0)& \textbf{41.57 (1.0)}& 17.75 (10.0)& 32.99 (5.0)& 16.52 (11.0)& 35.27 (4.0)& 28.72 (9.0)& 30.97 (7.0)& 0.00 (12.0)& 37.55 (2.0)\\
& D31& 21.19 (7.0)& 23.43 (6.0)& 13.09 (11.0)& 25.96 (2.0)& 13.61 (10.0)& 25.81 (3.0)& 25.28 (4.0)& \textbf{27.08 (1.0)}& 19.77 (9.0)& 20.12 (8.0)& 0.00 (12.0)& 23.73 (5.0)\\
& D32& 0.94 (9.0)& 1.23 (6.0)& 1.28 (5.0)& 1.94 (3.0)& \textbf{2.55 (1.0)}& 1.38 (4.0)& 1.00 (8.0)& 0.67 (10.0)& 0.47 (11.0)& 1.22 (7.0)& 0.00 (12.0)& 2.31 (2.0)\\
& D33& 0.70 (10.0)& 1.81 (8.0)& 3.55 (4.0)& 7.82 (2.0)& 3.11 (5.0)& 0.69 (11.0)& 0.58 (12.0)& 2.78 (6.0)& 1.07 (9.0)& 2.08 (7.0)& \textbf{9.67 (1.0)}& 7.31 (3.0)\\
& D34& 12.91 (5.0)& 1.10 (12.0)& 23.31 (3.0)& \textbf{31.87 (1.5)}& 7.36 (8.0)& 18.91 (4.0)& 12.34 (6.0)& 5.71 (9.0)& 12.08 (7.0)& 1.30 (11.0)& 2.47 (10.0)& \textbf{31.87 (1.5)}\\
& D35& 16.36 (4.0)& 0.09 (10.0)& 8.86 (6.0)& 17.37 (2.0)& 9.22 (5.0)& 6.13 (8.0)& 0.09 (10.0)& 17.11 (3.0)& 6.34 (7.0)& 0.09 (10.0)& 0.00 (12.0)& \textbf{17.65 (1.0)}\\
& D36& 0.40 (11.0)& 0.06 (12.0)& 5.13 (9.0)& 5.80 (8.0)& 20.04 (4.0)& 1.83 (10.0)& 20.17 (3.0)& \textbf{43.38 (1.0)}& 12.93 (5.0)& 35.47 (2.0)& 7.41 (7.0)& 12.91 (6.0)\\
& D37& 96.30 (10.0)& \textbf{100.00 (4.0)}& \textbf{100.00 (4.0)}& 99.78 (8.0)& 70.14 (11.0)& \textbf{100.00 (4.0)}& \textbf{100.00 (4.0)}& \textbf{100.00 (4.0)}& 96.89 (9.0)& \textbf{100.00 (4.0)}& 49.49 (12.0)& \textbf{100.00 (4.0)}\\
& D38& 0.15 (9.0)& 0.02 (11.5)& \textbf{19.84 (1.0)}& 1.54 (6.0)& 2.53 (5.0)& 8.54 (3.0)& 0.14 (10.0)& 0.56 (8.0)& 0.59 (7.0)& 0.02 (11.5)& 5.44 (4.0)& 11.43 (2.0)\\
& Average& 28.35 (9.42)& 61.09 (5.97)& 67.30 (4.54)& 56.54 (5.21)& 46.47 (7.54)& 69.34 (3.88)& 58.35 (6.62)& 57.67 (6.79)& 25.46 (9.92)& 50.80 (6.88)& 40.41 (8.34)& \textbf{72.95 (2.88)}\\
    \midrule
\multirow{39}{*}{ARI ($\uparrow$)} &  D1& 42.27 (9.0)& 24.47 (10.0)& 89.73 (5.0)& 98.32 (3.0)& 85.38 (6.0)& \textbf{100.00 (1.5)}& 1.28 (12.0)& 77.72 (7.0)& 21.86 (11.0)& 42.69 (8.0)& 96.67 (4.0)& \textbf{100.00 (1.5)}\\
& D2& 47.13 (12.0)& \textbf{100.00 (3.5)}& \textbf{100.00 (3.5)}& \textbf{100.00 (3.5)}& 91.79 (7.0)& \textbf{100.00 (3.5)}& \textbf{100.00 (3.5)}& 72.50 (10.0)& 55.82 (11.0)& 74.80 (9.0)& 75.65 (8.0)& \textbf{100.00 (3.5)}\\
& D3& 5.37 (12.0)& \textbf{100.00 (2.5)}& \textbf{100.00 (2.5)}& 30.16 (9.0)& 56.51 (7.0)& \textbf{100.00 (2.5)}& 77.45 (5.0)& 55.48 (8.0)& 7.05 (11.0)& 69.80 (6.0)& 14.58 (10.0)& \textbf{100.00 (2.5)}\\
& D4& -0.57 (12.0)& 45.94 (7.0)& \textbf{100.00 (3.0)}& 85.02 (6.0)& 8.47 (8.0)& \textbf{100.00 (3.0)}& \textbf{100.00 (3.0)}& \textbf{100.00 (3.0)}& 1.59 (10.0)& -0.48 (11.0)& 2.87 (9.0)& \textbf{100.00 (3.0)}\\
& D5& 54.11 (8.0)& 56.12 (7.0)& 93.81 (5.0)& \textbf{100.00 (2.5)}& 26.07 (12.0)& \textbf{100.00 (2.5)}& 86.65 (6.0)& 37.90 (9.0)& 34.42 (10.0)& 30.64 (11.0)& \textbf{100.00 (2.5)}& \textbf{100.00 (2.5)}\\
& D6& -0.17 (12.0)& \textbf{100.00 (3.5)}& \textbf{100.00 (3.5)}& 13.60 (10.0)& 25.92 (9.0)& \textbf{100.00 (3.5)}& \textbf{100.00 (3.5)}& 95.37 (7.0)& 1.07 (11.0)& \textbf{100.00 (3.5)}& 29.07 (8.0)& \textbf{100.00 (3.5)}\\
& D7& 65.81 (10.0)& 68.60 (9.0)& 86.88 (5.0)& 90.98 (3.0)& 69.16 (8.0)& 93.91 (2.0)& 76.44 (6.0)& 59.55 (11.0)& 56.44 (12.0)& 70.87 (7.0)& 90.34 (4.0)& \textbf{99.36 (1.0)}\\
& D8& 15.22 (8.0)& \textbf{100.00 (3.0)}& \textbf{100.00 (3.0)}& 39.00 (7.0)& 6.20 (10.0)& \textbf{100.00 (3.0)}& \textbf{100.00 (3.0)}& 97.60 (6.0)& 8.15 (9.0)& 0.00 (12.0)& 0.30 (11.0)& \textbf{100.00 (3.0)}\\
& D9& 74.59 (9.0)& 99.18 (4.0)& 92.63 (7.0)& 99.59 (2.5)& 59.11 (10.0)& 99.59 (2.5)& 37.12 (12.0)& 94.80 (6.0)& 47.62 (11.0)& 98.77 (5.0)& 76.85 (8.0)& \textbf{100.00 (1.0)}\\
& D10& 51.64 (11.0)& \textbf{100.00 (3.0)}& \textbf{100.00 (3.0)}& 62.63 (10.0)& 70.53 (9.0)& \textbf{100.00 (3.0)}& 95.56 (7.0)& 97.55 (6.0)& 43.63 (12.0)& 93.04 (8.0)& \textbf{100.00 (3.0)}& \textbf{100.00 (3.0)}\\
& D11& 13.74 (12.0)& \textbf{100.00 (4.0)}& \textbf{100.00 (4.0)}& 23.37 (10.0)& \textbf{100.00 (4.0)}& \textbf{100.00 (4.0)}& 70.98 (8.0)& 51.12 (9.0)& 16.36 (11.0)& \textbf{100.00 (4.0)}& \textbf{100.00 (4.0)}& \textbf{100.00 (4.0)}\\
& D12& 39.44 (12.0)& \textbf{100.00 (3.5)}& \textbf{100.00 (3.5)}& 74.76 (8.0)& 72.24 (9.0)& \textbf{100.00 (3.5)}& \textbf{100.00 (3.5)}& 93.07 (7.0)& 49.43 (11.0)& \textbf{100.00 (3.5)}& 54.45 (10.0)& \textbf{100.00 (3.5)}\\
& D13& 41.32 (10.0)& 76.26 (7.0)& \textbf{100.00 (2.0)}& 99.20 (5.0)& 39.89 (11.0)& \textbf{100.00 (2.0)}& 85.99 (6.0)& 54.32 (9.0)& 19.22 (12.0)& 65.58 (8.0)& 99.60 (4.0)& \textbf{100.00 (2.0)}\\
& D14& 9.40 (12.0)& \textbf{100.00 (3.5)}& \textbf{100.00 (3.5)}& 43.25 (9.0)& \textbf{100.00 (3.5)}& \textbf{100.00 (3.5)}& 59.11 (7.0)& 26.06 (10.0)& 12.78 (11.0)& \textbf{100.00 (3.5)}& 50.65 (8.0)& \textbf{100.00 (3.5)}\\
& D15& 48.82 (12.0)& \textbf{100.00 (3.0)}& \textbf{100.00 (3.0)}& 76.16 (8.0)& 67.42 (10.0)& \textbf{100.00 (3.0)}& \textbf{100.00 (3.0)}& 97.62 (6.0)& 50.48 (11.0)& 96.31 (7.0)& 70.91 (9.0)& \textbf{100.00 (3.0)}\\
& D16& 39.86 (8.0)& 66.57 (7.0)& \textbf{100.00 (3.0)}& 78.18 (6.0)& 13.99 (10.0)& \textbf{100.00 (3.0)}& \textbf{100.00 (3.0)}& \textbf{100.00 (3.0)}& 7.23 (11.0)& 0.00 (12.0)& 33.22 (9.0)& \textbf{100.00 (3.0)}\\
& D17& -0.30 (12.0)& \textbf{100.00 (3.5)}& \textbf{100.00 (3.5)}& 70.84 (7.0)& 27.13 (10.0)& \textbf{100.00 (3.5)}& \textbf{100.00 (3.5)}& \textbf{100.00 (3.5)}& 1.33 (11.0)& 33.14 (9.0)& 38.77 (8.0)& \textbf{100.00 (3.5)}\\
& D18& -0.10 (12.0)& \textbf{100.00 (5.5)}& \textbf{100.00 (5.5)}& \textbf{100.00 (5.5)}& \textbf{100.00 (5.5)}& \textbf{100.00 (5.5)}& \textbf{100.00 (5.5)}& \textbf{100.00 (5.5)}& 16.53 (11.0)& \textbf{100.00 (5.5)}& \textbf{100.00 (5.5)}& \textbf{100.00 (5.5)}\\
& D19& 0.75 (12.0)& 11.37 (6.0)& 2.17 (11.0)& 19.66 (2.0)& 4.89 (9.0)& 19.30 (3.0)& 6.72 (7.0)& 2.38 (10.0)& 5.99 (8.0)& 15.00 (4.0)& 13.79 (5.0)& \textbf{84.62 (1.0)}\\
& D20& 27.33 (11.0)& \textbf{100.00 (4.5)}& \textbf{100.00 (4.5)}& \textbf{100.00 (4.5)}& \textbf{100.00 (4.5)}& \textbf{100.00 (4.5)}& 69.91 (9.0)& 37.91 (10.0)& 1.31 (12.0)& \textbf{100.00 (4.5)}& \textbf{100.00 (4.5)}& \textbf{100.00 (4.5)}\\
& D21& 3.62 (12.0)& \textbf{100.00 (4.0)}& \textbf{100.00 (4.0)}& \textbf{100.00 (4.0)}& 14.51 (10.0)& \textbf{100.00 (4.0)}& \textbf{100.00 (4.0)}& \textbf{100.00 (4.0)}& 4.96 (11.0)& 19.20 (8.0)& 15.28 (9.0)& \textbf{100.00 (4.0)}\\
& D22& 0.04 (11.0)& \textbf{100.00 (3.5)}& \textbf{100.00 (3.5)}& 46.71 (9.0)& \textbf{100.00 (3.5)}& \textbf{100.00 (3.5)}& 81.75 (8.0)& 95.31 (7.0)& 7.13 (10.0)& \textbf{100.00 (3.5)}& 0.00 (12.0)& \textbf{100.00 (3.5)}\\
& D23& 36.31 (11.0)& 83.78 (3.0)& 62.90 (7.0)& 81.74 (5.0)& 40.31 (10.0)& \textbf{100.00 (1.5)}& 83.50 (4.0)& 74.66 (6.0)& 35.91 (12.0)& 55.22 (8.0)& 45.91 (9.0)& \textbf{100.00 (1.5)}\\
& D24& 41.43 (11.0)& \textbf{100.00 (4.0)}& \textbf{100.00 (4.0)}& 56.07 (9.0)& \textbf{100.00 (4.0)}& \textbf{100.00 (4.0)}& \textbf{100.00 (4.0)}& 99.93 (8.0)& 24.68 (12.0)& \textbf{100.00 (4.0)}& 55.82 (10.0)& \textbf{100.00 (4.0)}\\
& D25& \textbf{36.31 (1.0)}& 7.16 (7.0)& 0.00 (11.5)& 16.23 (3.0)& 8.98 (5.0)& 9.15 (4.0)& 6.26 (8.0)& 4.87 (9.0)& 7.19 (6.0)& 2.96 (10.0)& 0.00 (11.5)& 32.44 (2.0)\\
& D26& 26.89 (3.0)& 7.32 (9.0)& 26.95 (2.0)& 14.37 (4.0)& 10.07 (6.0)& 8.71 (8.0)& 1.37 (11.0)& 9.92 (7.0)& 1.50 (10.0)& 12.65 (5.0)& 0.00 (12.0)& \textbf{46.16 (1.0)}\\
& D27& 58.26 (5.0)& 44.71 (9.0)& 56.81 (6.0)& 75.76 (2.0)& 54.37 (7.0)& \textbf{90.38 (1.0)}& 46.99 (8.0)& 44.12 (10.0)& 41.05 (11.0)& 67.37 (3.0)& 0.00 (12.0)& 65.37 (4.0)\\
& D28& -0.12 (12.0)& 3.29 (3.5)& 0.79 (6.0)& \textbf{6.15 (1.0)}& 0.00 (10.5)& 2.30 (5.0)& 0.09 (9.0)& 0.58 (7.0)& 0.13 (8.0)& 5.16 (2.0)& 0.00 (10.5)& 3.29 (3.5)\\
& D29& 1.61 (5.0)& -0.46 (8.0)& 2.10 (4.0)& \textbf{19.46 (1.0)}& 0.84 (6.0)& 14.46 (2.0)& -0.71 (9.0)& -1.99 (12.0)& -0.84 (10.0)& -1.72 (11.0)& 0.00 (7.0)& 7.69 (3.0)\\
& D30& 17.75 (6.0)& 12.51 (9.0)& 23.89 (3.0)& \textbf{28.93 (1.0)}& 7.68 (10.0)& 26.90 (2.0)& 5.48 (11.0)& 18.82 (5.0)& 15.51 (8.0)& 15.64 (7.0)& 0.00 (12.0)& 23.80 (4.0)\\
& D31& 1.67 (8.0)& -6.19 (12.0)& 5.44 (4.0)& \textbf{13.56 (1.0)}& 4.17 (6.0)& 5.68 (3.0)& 3.77 (7.0)& 5.12 (5.0)& -1.74 (10.0)& -4.84 (11.0)& 0.00 (9.0)& 9.09 (2.0)\\
& D32& -1.79 (12.0)& 0.37 (4.0)& 0.59 (3.0)& 0.27 (6.0)& 1.45 (2.0)& 0.09 (7.0)& -0.62 (11.0)& -0.02 (9.0)& -0.41 (10.0)& 0.34 (5.0)& 0.00 (8.0)& \textbf{3.38 (1.0)}\\
& D33& 4.29 (4.0)& -2.69 (11.0)& 2.71 (7.0)& 6.77 (2.0)& 1.29 (10.0)& 2.12 (8.0)& 1.31 (9.0)& -3.05 (12.0)& 2.75 (6.0)& 3.11 (5.0)& \textbf{12.20 (1.0)}& 6.51 (3.0)\\
& D34& -2.19 (9.0)& -2.21 (10.0)& 30.08 (3.0)& \textbf{44.12 (1.5)}& 20.14 (5.0)& 27.39 (4.0)& 12.18 (8.0)& 15.22 (6.0)& 13.02 (7.0)& -2.90 (11.0)& -7.13 (12.0)& \textbf{44.12 (1.5)}\\
& D35& 13.59 (4.0)& 0.00 (10.5)& 2.83 (6.0)& \textbf{18.85 (1.0)}& 4.87 (5.0)& 1.69 (8.0)& 0.00 (10.5)& 13.62 (3.0)& 2.51 (7.0)& 0.00 (10.5)& 0.00 (10.5)& 18.24 (2.0)\\
& D36& 0.49 (11.0)& 0.03 (12.0)& 9.41 (8.0)& 11.42 (6.0)& 23.36 (3.0)& 1.30 (10.0)& 22.54 (4.0)& \textbf{46.27 (1.0)}& 15.29 (5.0)& 40.66 (2.0)& 10.07 (7.0)& 7.64 (9.0)\\
& D37& 86.49 (10.0)& \textbf{100.00 (4.0)}& \textbf{100.00 (4.0)}& 98.75 (8.0)& 19.38 (11.0)& \textbf{100.00 (4.0)}& \textbf{100.00 (4.0)}& \textbf{100.00 (4.0)}& 87.95 (9.0)& \textbf{100.00 (4.0)}& 6.99 (12.0)& \textbf{100.00 (4.0)}\\
& D38& 0.63 (7.0)& 0.01 (11.5)& \textbf{24.90 (1.0)}& 4.23 (5.0)& 0.51 (8.0)& 6.05 (3.0)& 0.08 (10.0)& 1.96 (6.0)& 0.43 (9.0)& 0.01 (11.5)& 4.82 (4.0)& 12.62 (2.0)\\
& Average& 23.71 (9.42)& 57.79 (6.18)& 66.17 (4.50)& 53.90 (5.03)& 40.44 (7.49)& 68.66 (3.80)& 56.08 (6.74)& 54.64 (7.03)& 18.82 (9.95)& 47.45 (6.92)& 36.62 (7.97)& \textbf{72.75 (2.97)}\\
\midrule  
    CO ($\downarrow$) & All & 0.12 (1) &	7.36 (8) & 1.08 (2) &	4.55 (7) &	808.39 (12) & 1.54 (3) &	3.49 (6) &	69.79 (11) &	1.67 (4) &	7.55 (9) &	18.16 (10) &	3.30 (5) \\
    \bottomrule
\end{tabular}
    \end{threeparttable}
\end{table*}

\begin{figure*}[!htbp]
  \centering
  \includegraphics[width=\linewidth]{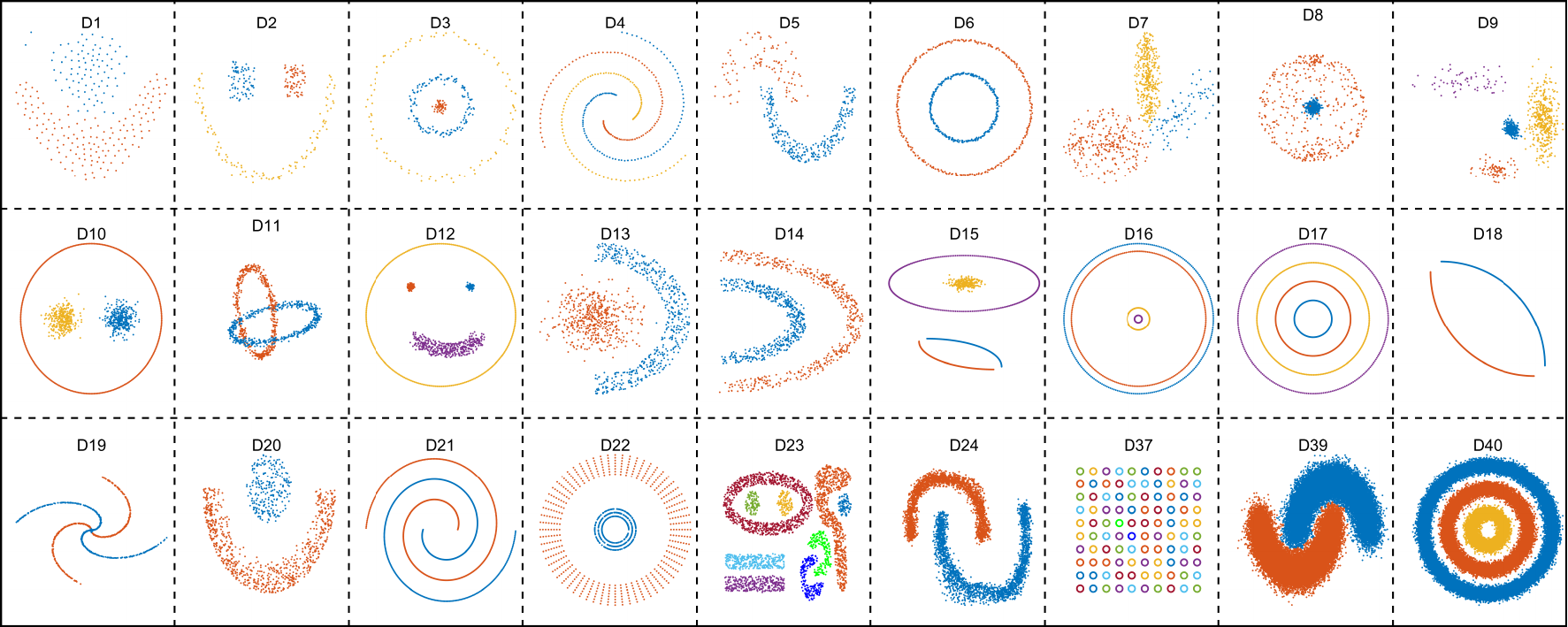}
  \caption{Clustering results of LGBQPC on 27 synthetic datasets.}
\label{Figure:Clustering-Results-LGBQPC-Synthetic-Datasets}
\end{figure*}

\begin{figure}[!htbp]
  \centering
  \subfigure[]{\includegraphics[width=0.45\linewidth]{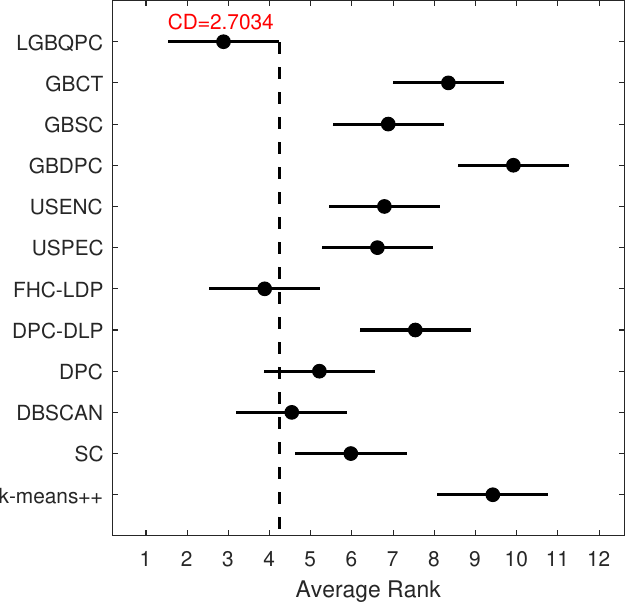}}
  \subfigure[]{\includegraphics[width=0.45\linewidth]{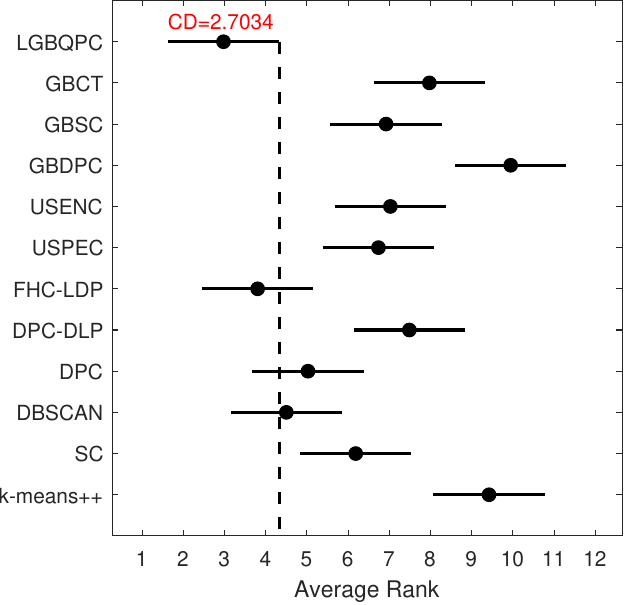}}
  \caption{Results of Nemenyi test on datasets D1-D38. (a) NMI. (b) ARI.}
  \label{Figure:Nemenyi's Test}
\end{figure}

The experimental results are summarized in Tables \ref{Table:Experimental Results on Datasets D1-D38} and \ref{Table:Experimental Results on Datasets D39-D40}, along with Fig. \ref{Figure:Clustering-Results-LGBQPC-Synthetic-Datasets}. Specifically, Table \ref{Table:Experimental Results on Datasets D1-D38} presents the performance of 12 clustering algorithms on the standard-sized datasets D1-D38, whereas Table \ref{Table:Experimental Results on Datasets D39-D40} evaluates nine scalable clustering algorithms on the large-scale datasets D39-D40. Furthermore, Fig. \ref{Figure:Clustering-Results-LGBQPC-Synthetic-Datasets} visualizes the clustering results of LGBQPC across all synthetic datasets.

As shown in Table \ref{Table:Experimental Results on Datasets D1-D38}, LGBQPC demonstrates superior performance compared to the other methods, achieving the highest NMI on 29 datasets and the highest ARI on 28 datasets. Additionally, it achieves the best average NMI and ARI across all standard-sized datasets. In terms of CO, LGBQPC ranks fifth among the 12 compared methods, highlighting a favorable balance between accuracy and efficiency. Figure \ref{Figure:Clustering-Results-LGBQPC-Synthetic-Datasets} further confirms  its effectiveness in handling complex manifold structures and non-uniform density distributions (particularly evident in datasets D5 and D7-D9). 

Following the Friedman test outlined in Section \ref{Section4.1}, the results from Table \ref{Table:Experimental Results on Datasets D1-D38} yield test statistics of $\tau_{F}=20.31$ for NMI and $\tau_{F}=19.93$ for ARI. Since both values exceed the critical threshold $F_{0.05}(11,407)=1.81$, the differences among the 12 clustering algorithms are statistically significant at the 0.05 level. Furthermore,  the Nemenyi test results for both NMI and ARI, as presented in Fig. \ref{Figure:Nemenyi's Test}, indicate that LGBQPC significantly outperforms eight of the compared clustering algorithms, particularly three GB-based algorithms: GBSC, GBDPC, and GBCT.

On the other hand, Table \ref{Table:Experimental Results on Datasets D39-D40} demonstrates that LGBQPC attains superior clustering performance on the large-scale datasets D39 and D40, ranking first and second, respectively. Simultaneously, it maintains acceptable computational efficiency ranking seventh and fifth among the nine scalable algorithms. These results suggest that LGBQPC is well-suited for large-scale data analysis. 

In summary, LGBQPC achieves statistically significant improvements over GBDPC, establishing itself an efficient, effective, and scalable clustering method capable of reliably addressing datasets with complex manifold structures or non-uniform density distributions.

\begin{table}[!htbp]
  \centering
  \fontsize{6}{6}\selectfont
  \renewcommand\tabcolsep{4pt}
  \begin{threeparttable}
  \caption{Experimental Results on Datasets D39-D40}
  \label{Table:Experimental Results on Datasets D39-D40}
  \begin{tabular}{lllllll}
    \toprule
    \multirow{2}{*}{Algorithm} & \multicolumn{3}{c}{D39}                                     & \multicolumn{3}{c}{D40}                                     \\
    \cmidrule(lr){2-4}\cmidrule(lr){5-7}
                                & NMI ($\uparrow$)& ARI ($\uparrow$)& CO ($\downarrow$)& NMI ($\uparrow$)& ARI ($\uparrow$)& CO ($\downarrow$)\\
    \midrule
    $k$-means++ & 44.650 (9)          & 55.182 (9)          & \textbf{0.036 (1)} & 0.002 (9)           & 0.001 (9)           & \textbf{0.154 (1)} \\
DBSCAN  & 78.364 (6)          & 89.613 (6)          & 1.467 (6)          & 99.658 (6)          & 99.882 (6)          & 1.607 (6)          \\
FHC-LDP & 96.909 (2)    & 98.731 (2)    & 0.217 (2)    & 99.867 (3)          & 99.958 (3)          & 2.162 (7)          \\
USPEC   & 95.473 (5)          & 98.013 (5)          & 0.543 (3)          & 99.865 (4)          & 99.962 (4)          & 0.513 (2)    \\
USENC   & 96.154 (4)          & 98.130 (4)          & 12.375 (9)         & \textbf{99.904 (1)} & \textbf{99.976 (1)} & 13.840 (9)         \\
GBDPC   & 75.252 (7)          & 75.396 (7)          & 0.679 (4)          & 13.818 (7)          & 0.137 (7)           & 0.793 (3)          \\
GBSC    & 46.573 (8)          & 57.163 (8)          & 1.256 (5)          & 4.307 (8)           & 0.002 (8)           & 1.293 (4)          \\
GBCT    & 96.317 (3)          & 98.425 (3)          & 6.837 (8)          & 99.856 (5)          & 99.961 (5)          & 9.163 (8)          \\
LGBQPC (Ours)  & \textbf{97.040 (1)} & \textbf{98.797 (1)} & 4.688 (7)          & 99.881 (2)    & 99.964 (2)    & 1.599 (5) \\
    \bottomrule
    \end{tabular}
    \end{threeparttable}
\end{table}

\subsection{Parameter Analysis}
\begin{figure}[htbp]
  \centering
  \subfigure[]{\includegraphics[width=0.45\linewidth]{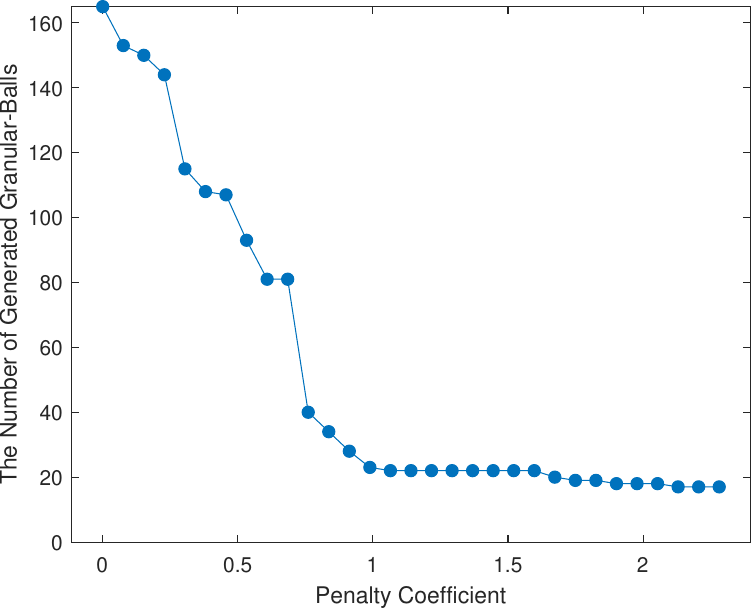}}
  \subfigure[]{\includegraphics[width=0.45\linewidth]{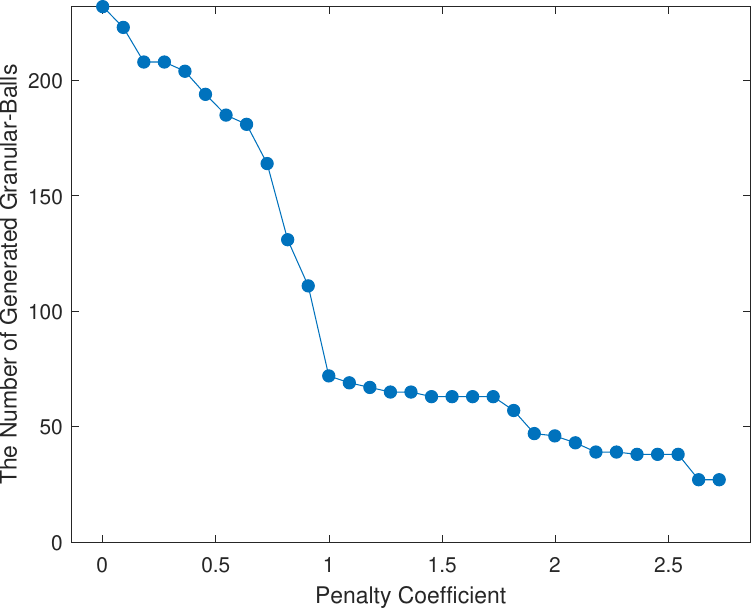}}
  \subfigure[]{\includegraphics[width=0.45\linewidth]{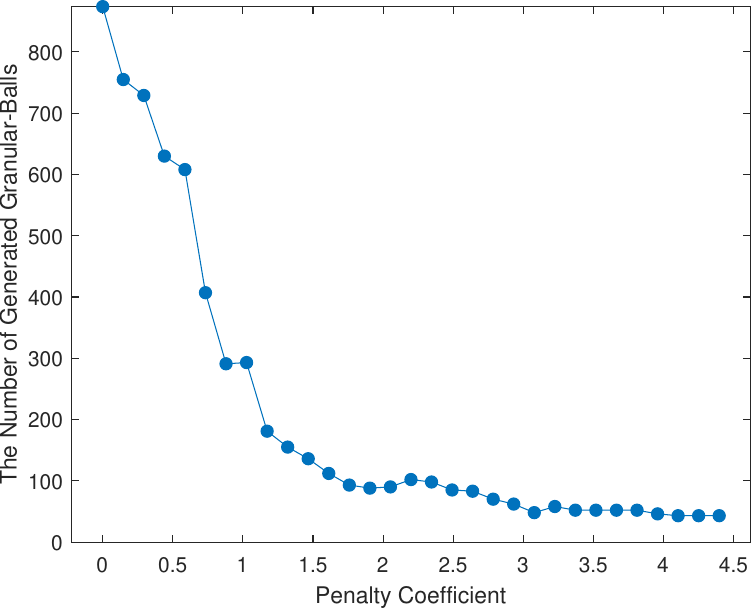}}
  \subfigure[]{\includegraphics[width=0.45\linewidth]{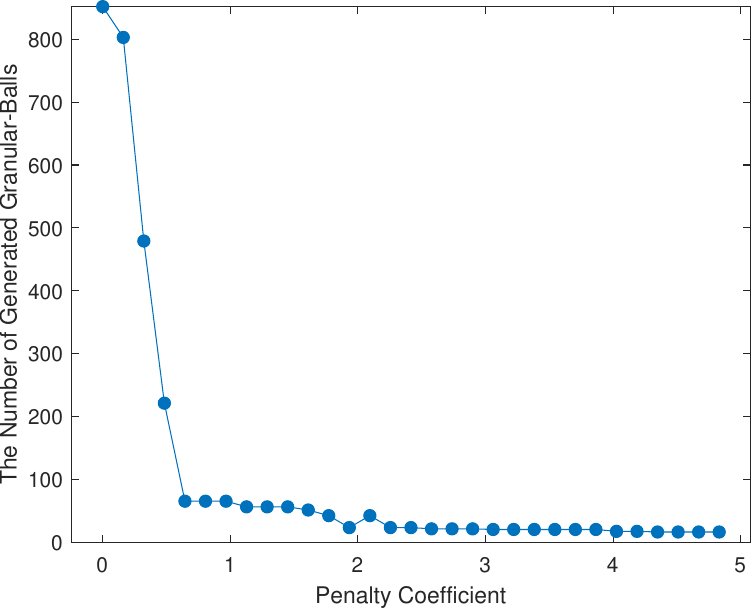}}
  \caption{Number of GBs generated by GB-POJG+ under varying penalty coefficients. (a)-(d) D32-D35.}
  \label{Figure: Performance of GB-POJG+ with respect to different values}
\end{figure}

\begin{figure}[htbp]
  \centering
  \subfigure[]{\includegraphics[width=0.45\linewidth]{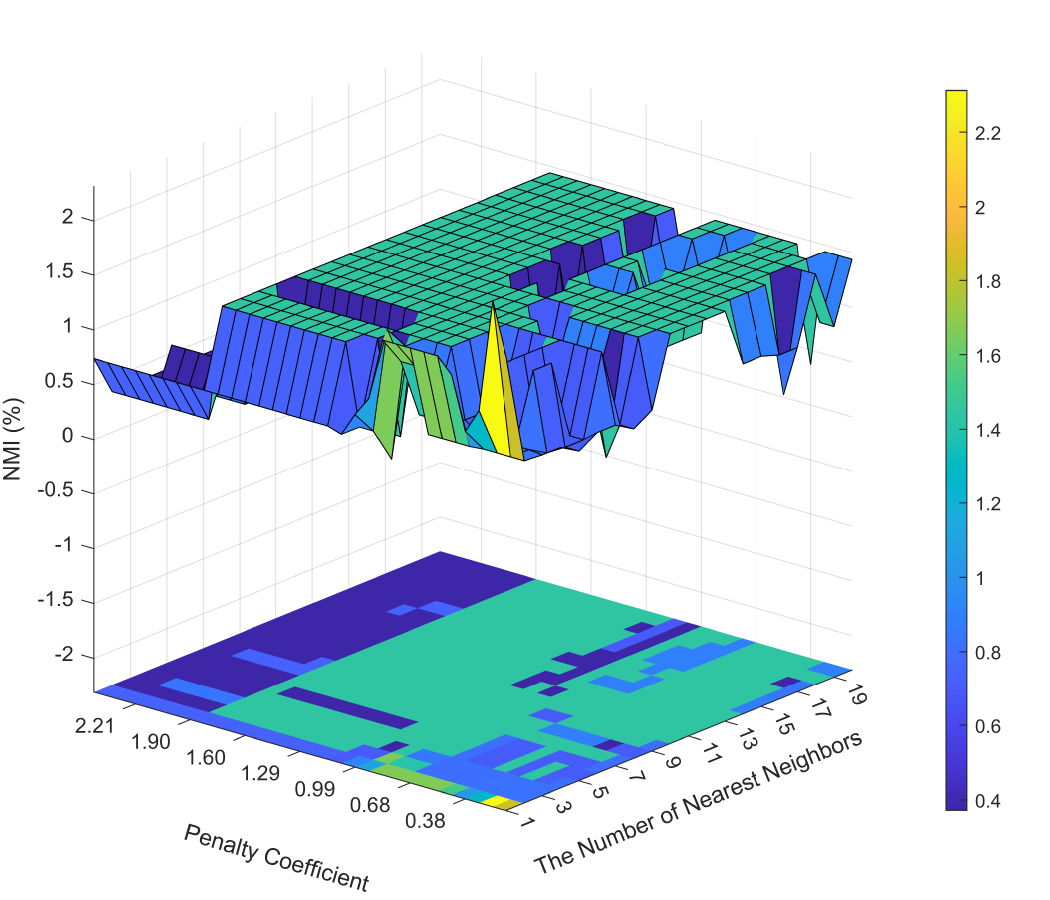}}
  \subfigure[]{\includegraphics[width=0.45\linewidth]{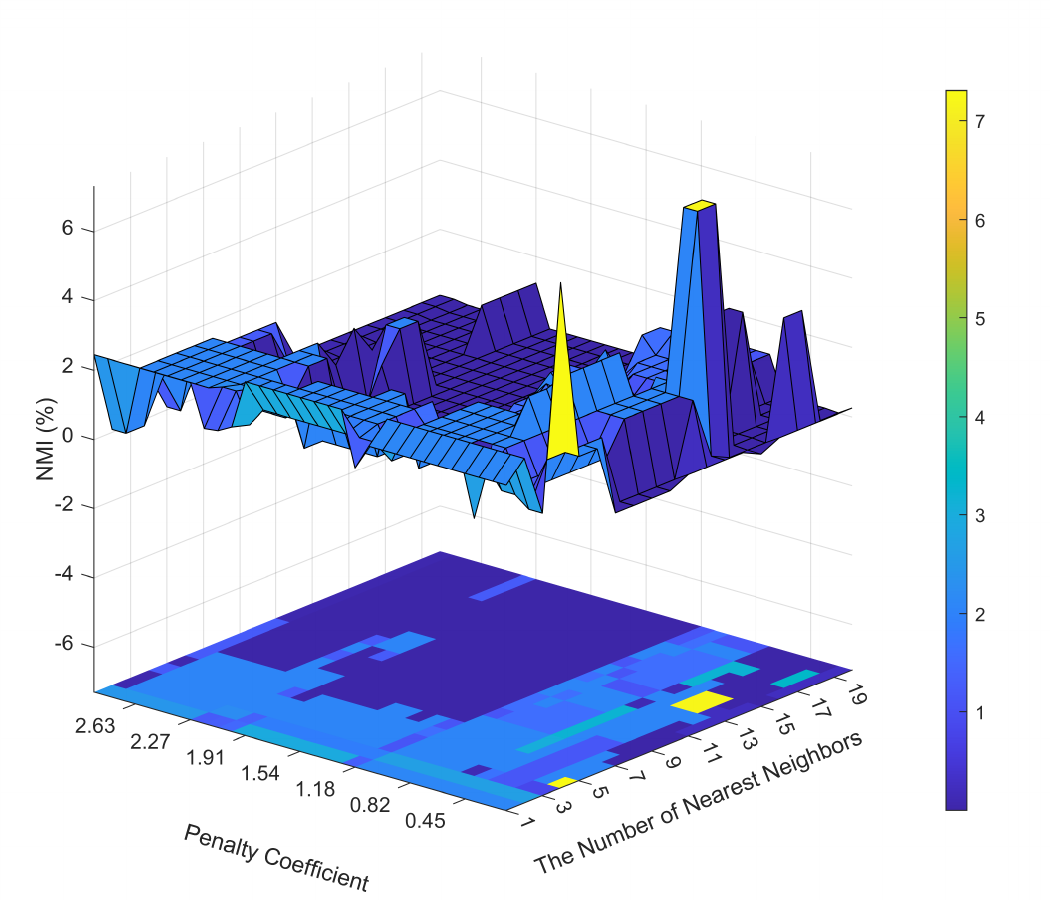}}
  \subfigure[]{\includegraphics[width=0.45\linewidth]{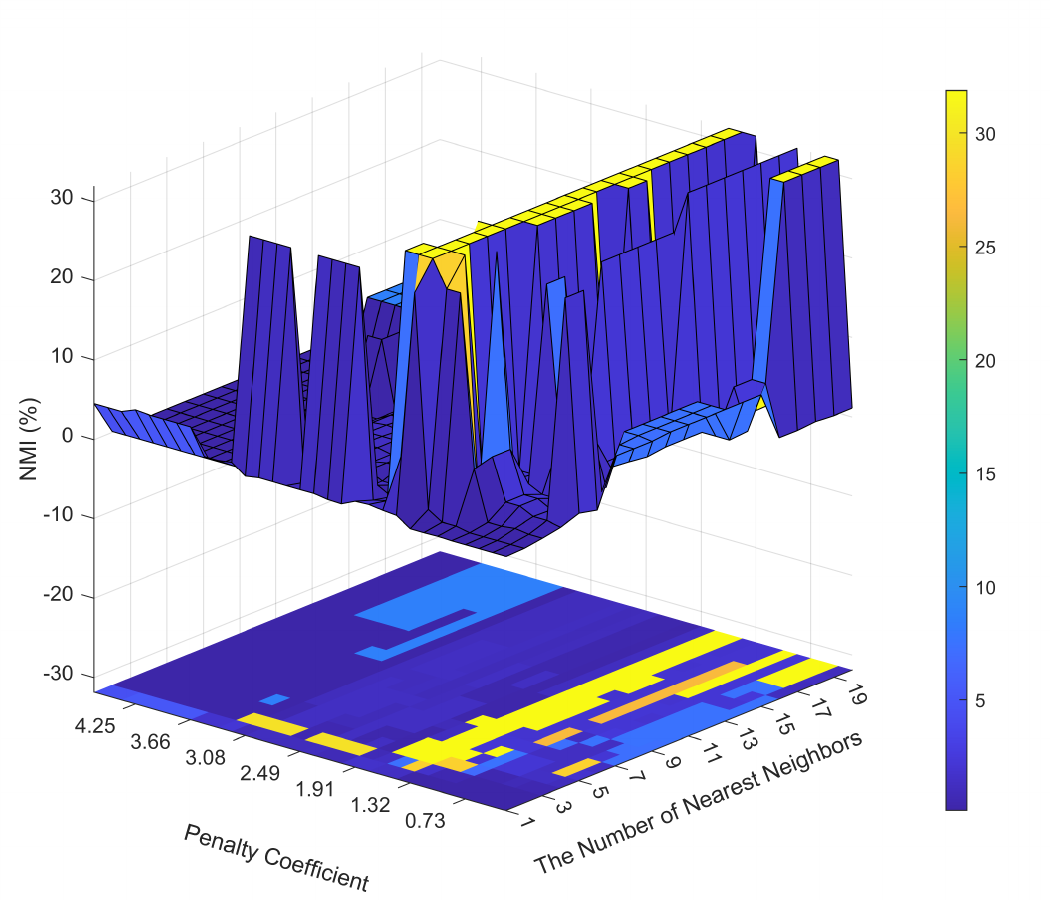}}
  \subfigure[]{\includegraphics[width=0.45\linewidth]{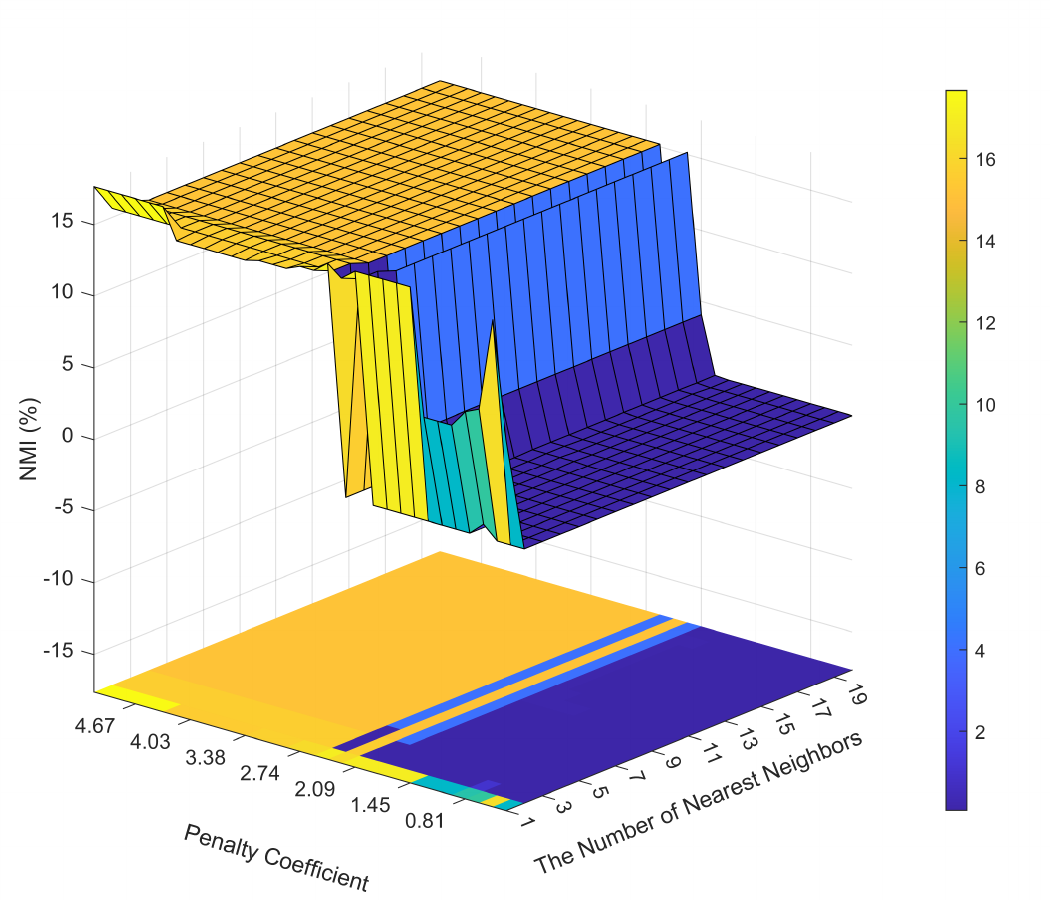}}
  \caption{Performance of LGBQPC under varying parameter configurations. (a)-(d) D32-D35.}
  \label{Figure: Performance of LGBQPC with respect to different values}
\end{figure}

This section systematically evaluates the performance of LGBQPC under varying parameter configurations. The experiments are conducted on four representative  datasets (D32-D35). The penalty coefficient ($\lambda$) is varied from 0 to $0.3\sqrt[3]{n}$ in increments of $0.01\sqrt[3]{n}$, while the number of nearest neighbors ($k$) is adjusted from 1 to 20 in steps of 1. The experimental results are presented in Figs. \ref{Figure: Performance of GB-POJG+ with respect to different values} and \ref{Figure: Performance of LGBQPC with respect to different values}, respectively. Specifically, Fig. \ref{Figure: Performance of GB-POJG+ with respect to different values} illustrates the number of generated GBs across different penalty coefficients, and Fig. \ref{Figure: Performance of LGBQPC with respect to different values} evaluates the clustering performance of LGBQPC under different parameter configurations using NMI. 

As illustrated in Fig. \ref{Figure: Performance of GB-POJG+ with respect to different values}, the number of generated GBs follows a consistent pattern across all selected datasets, decreasing monotonically as the penalty coefficient increases. This indicates that the number of generated GBs can be efficiently controlled within an acceptable range by tuning the penalty coefficient in GB-POJG+.

Furthermore, Fig. \ref{Figure: Performance of LGBQPC with respect to different values} reveals that the best performance of LGBQPC is consistently achieved when a non-zero penalty coefficient is set across all selected datasets. This observation highlights the importance of integrating a penalty coefficient into the objective function of GB-POJG+.  These findings suggest that generating GBs at coarser granularity, rather than the finest, enhances the performance of LGBQPC. Additionally, LGBQPC exhibits insensitivity to small variations in the number of nearest neighbors (used for constructing the GB $k$-NN graph), demonstrating relatively robust performance.

\subsection{Ablation Studies}\label{Section:Ablation-Study}
This section conducts a systematic analysis of the contributions of individual components within LGBQPC through comparison with eight ablated variants. These  variants are configured as follows:
\begin{itemize}
    \item LGBQPC-v1, -v2, and -v3: Adopt definitions of abnormal GBs from \cite{JiaZihang2025TCYB}, \cite{XieJiang2023TKDE}, and \cite{XieJiang2024ICDE}, respectively.
    \item LGBQPC-v4: Identifies a GB as abnormal if it satisfies either the criteria from \cite{XieJiang2023TKDE} or \cite{JiaZihang2025TCYB}.
    \item LGBQPC-v5: Ensures each leaf node in the GB-based binary tree is sufficiently small but omits anomaly detection and subsequent complete division.
   \item LGBQPC-v6: Calculates GB density using Eq. (\ref{Equation:Granular-Ball-Density-GBDPC}).
   \item LGBQPC-v7: Estimates GB density based on Definition \ref{Definition:Quality of Granular-Ball} and Eq. (\ref{Equation:Increasing-Decreasing-Functions}), without considering local structural information of GBs.
\item  LGBQPC-v8: Employs Eq. (\ref{Equation:Distance-Between-Granular-Balls}) as the distance metric for GBs, replacing the geodesic distance in Definition \ref{Definition:Geodesic-Distance-Granular-Balls}. 
\end{itemize}

The experiments are performed on all datasets D1-D40. Notably, for each dataset,  LGBQPC and its variants share the same parameter settings, where the optimal configuration for LGBQPC is uniformly applied. The results are summarized in Table \ref{Table:Experimental Results on Ablation Studies}. From Table \ref{Table:Experimental Results on Ablation Studies}, the following insights can be drawn:
\begin{enumerate}
\item [(1)] Comparing LGBQPC with its variants v1–v3  indicates that the definition of abnormal GBs presented in Definition \ref{Definition:Abnormal-Granular-Balls-Sigma} is more appropriate for LGBQPC than those proposed in \cite{XieJiang2023TKDE,XieJiang2024ICDE,JiaZihang2025TCYB}. Moreover, the performance of LGBQPC-v4 not only demonstrates the necessity of defining abnormal GBs using both average and maximum radii but also underscores the superiority of adaptive thresholds determined by mean and standard deviation over fixed ones.
\item [(2)] The inferior results of LGBQPC-v5 demonstrate that the detection and complete division of abnormal leaf nodes during GB-based binary tree construction is essential for generating GBs with enhanced clustering suitability. In the other words, this mechanism effectively prevents the generation of GBs either located at decision boundaries or containing noise instances.
\item [(3)] LGBQPC outperforms its variants v6 and v7, demonstrating that the relative quality metric in Definition \ref{Definition:Relative-Quality-Granular-Balls} serves as a more effective density estimator for GBs. Avoiding density overestimation and incorporating local neighborhood information of GBs are two effective strategies for GB density estimation.
\item [(4)] The experimental evaluation of LGBQPC-v8 reveals that adopting geodesic distance enhances LGBQPC's ability to identify complex manifold structures, thereby effectively improving its clustering performance.
\end{enumerate}

In summary, LGBQPC achieves superior clustering performance through effective improvements in: (1) abnormal GB definition, (2) GB-based binary tree construction, (3) GB density estimation, and (4) GB distance metric.

\begin{table}[!htbp]
  \centering
  \fontsize{6}{6}\selectfont
  \renewcommand\tabcolsep{10pt}
  \begin{threeparttable}
  \caption{Results of Ablation Studies on Datasets D1-D40}
  \label{Table:Experimental Results on Ablation Studies}
  \begin{tabular}{llll}
    \toprule
    Algorithm & Average NMI ($\uparrow$)    & Average ARI ($\uparrow$)    & All CO ($\downarrow$)        \\
    \midrule
    LGBQPC (Ours)  & \textbf{74.22} & \textbf{74.08} & 9.59         \\
    LGBQPC-v1 & 58.45          & 55.86          & \textbf{4.73} \\
    LGBQPC-v2 & 64.11          & 62.42          & 7.69          \\
    LGBQPC-v3 & 58.23          & 54.11          & 20.75         \\
    LGBQPC-v4 & 67.75          & 66.22          & 7.55          \\
    LGBQPC-v5 & 59.12          & 56.6           & 5.03          \\
    LGBQPC-v6 & 61.58          & 59.78          & 10.98         \\
    LGBQPC-v7 & 68.57          & 67.15          & 10.04         \\
    LGBQPC-v8 & 26.17          & 19.45          & 8.55\\         
    \bottomrule
    \end{tabular}
    \end{threeparttable}
\end{table}

\section{Conclusion}\label{Section5}
This paper proposes a novel clustering algorithm, termed LGBQPC, which significantly improves the GBDPC algorithm  through enhancements from both GB generation and clustering perspectives. Specifically, an advanced GB generation method, GB-POJG+, is developed building upon the GB-POJG framework. The proposed GB-POJG+ requires only a single parameter, namely the penalty coefficient, to ensure the generation of high-quality GBs while keeping their quantity within an acceptable range. Furthermore, to address the limitations of GBDPC in handling non-uniform density distributions and complex manifold structures, two key innovations are introduced based on the GB $k$-NN graph: (1) the use of relative quality as an enhanced GB density estimator, and (2) the adoption of geodesic distance as an improved GB distance metric. These contributions collectively establish LGBQPC as an efficient, effective, and scalable clustering algorithm.

In future work, given the challenges of applying current GB computing methods to handling high-dimensional data \cite{JiaZihang2025TCYB}, it is worthwhile to further investigate enhanced GB generation strategies and develop corresponding clustering methods for unlabeled high-dimensional data. Additionally, exploring effective strategies to automatically determine the optimal number of nearest neighbors for GBs in LGBQPC may offer further improvements in clustering performance  \cite{ZhuQingsheng2016PRL}.

\ifCLASSOPTIONcaptionsoff
  \newpage
\fi

% trigger a \newpage just before the given reference
% number - used to balance the columns on the last page
% adjust value as needed - may need to be readjusted if
% the document is modified later
%\IEEEtriggeratref{8}
% The "triggered" command can be changed if desired:
%\IEEEtriggercmd{\enlargethispage{-5in}}

% references section

% can use a bibliography generated by BibTeX as a .bbl file
% BibTeX documentation can be easily obtained at:
% http://mirror.ctan.org/biblio/bibtex/contrib/doc/
% The IEEEtran BibTeX style support page is at:
% http://www.michaelshell.org/tex/ieeetran/bibtex/
\bibliography{reference}

% Generated by IEEEtran.bst, version: 1.14 (2015/08/26)
\begin{thebibliography}{10}
\providecommand{\url}[1]{#1}
\csname url@samestyle\endcsname
\providecommand{\newblock}{\relax}
\providecommand{\bibinfo}[2]{#2}
\providecommand{\BIBentrySTDinterwordspacing}{\spaceskip=0pt\relax}
\providecommand{\BIBentryALTinterwordstretchfactor}{4}
\providecommand{\BIBentryALTinterwordspacing}{\spaceskip=\fontdimen2\font plus
\BIBentryALTinterwordstretchfactor\fontdimen3\font minus
  \fontdimen4\font\relax}
\providecommand{\BIBforeignlanguage}[2]{{%
\expandafter\ifx\csname l@#1\endcsname\relax
\typeout{** WARNING: IEEEtran.bst: No hyphenation pattern has been}%
\typeout{** loaded for the language `#1'. Using the pattern for}%
\typeout{** the default language instead.}%
\else
\language=\csname l@#1\endcsname
\fi
#2}}
\providecommand{\BIBdecl}{\relax}
\BIBdecl

\bibitem{XuRui2005TNNLS}
R.~Xu and D.~C. Wunsch, ``Survey of clustering algorithms,'' \emph{IEEE
  Transactions on Neural Networks}, vol.~16, no.~3, pp. 645--678, 2005.

\bibitem{Arthur2007ACM-SIAM}
D.~Arthur and S.~Vassilvitskii, ``K-means++: {T}he advantages of careful
  seeding,'' in \emph{Proceedings of the 18th Annual ACM-SIAM Symposium on
  Discrete Algorithms}, 2007, pp. 1027 -- 1035.

\bibitem{VonLuxburg2007StatisticsComputing}
U.~Von~Luxburg, ``A tutorial on spectral clustering,'' \emph{Statistics and
  Computing}, vol.~17, no.~4, pp. 395 -- 416, 2007.

\bibitem{Ester1996KDD}
M.~Ester, H.~P. Kriegel, J.~Sander, and X.~Xu, ``A density-based algorithm for
  discovering clusters a density-based algorithm for discovering clusters in
  large spatial databases with noise,'' in \emph{Proceedings of the 2nd
  International Conference on Knowledge Discovery and Data Mining (KDD)}, 1996,
  pp. 226--231.

\bibitem{Rodriguez2014Science}
A.~Rodriguez and A.~Laio, ``Clustering by fast search and find of density
  peaks,'' \emph{Science}, vol. 344, no. 6191, pp. 1492--1496, 2014.

\bibitem{WangYizhang2024ESWA}
Y.~Wang, J.~Qian, M.~Hassan, X.~Zhang, T.~Zhang, C.~Yang, X.~Zhou, and F.~Jia,
  ``Density peak clustering algorithms: {A} review on the decade 2014-2023,''
  \emph{Expert Systems with Applications}, vol. 238, 2024, {A}rt. no. 121860.

\bibitem{HouJian2020TII}
J.~Hou and A.~Zhang, ``Enhancing density peak clustering via density
  normalization,'' \emph{IEEE Transactions on Industrial Informatics}, vol.~16,
  no.~4, pp. 2477--2485, 2020.

\bibitem{FangXintong2023TII}
X.~Fang, Z.~Xu, H.~Ji, B.~Wang, and Z.~Huang, ``A grid-based density peaks
  clustering algorithm,'' \emph{IEEE Transactions on Industrial Informatics},
  vol.~19, no.~4, pp. 5476--5484, 2023.

\bibitem{TongWuning2023TKDE}
W.~Tong, Y.~Wang, and D.~Liu, ``An adaptive clustering algorithm based on
  local-density peaks for imbalanced data without parameters,'' \emph{IEEE
  Transactions on Knowledge and Data Engineering}, vol.~35, no.~4, pp.
  3419--3432, 2023.

\bibitem{dErrico2021INS}
M.~d'Errico, E.~Facco, A.~Laio, and A.~Rodriguez, ``Automatic topography of
  high-dimensional data sets by non-parametric density peak clustering,''
  \emph{Information Sciences}, vol. 560, pp. 476--492, 2021.

\bibitem{DuMingjing2018IJMLC}
M.~Du, S.~Ding, X.~Xu, and Y.~Xue, ``Density peaks clustering using geodesic
  distances,'' \emph{International Journal of Machine Learning and
  Cybernetics}, vol.~9, pp. 1335--1349, 2017.

\bibitem{DingShifei2017KBS}
S.~Ding, M.~Du, T.~Sun, X.~Xu, and Y.~Xue, ``An entropy-based density peaks
  clustering algorithm for mixed type data employing fuzzy neighborhood,''
  \emph{Knowledge-Based Systems}, vol. 133, pp. 294--313, 2017.

\bibitem{ChengDongdong2024TNNLS}
D.~Cheng, Y.~Li, S.~Xia, G.~Wang, J.~Huang, and S.~Zhang, ``A fast
  granular-ball-based density peaks clustering algorithm for large-scale
  data,'' \emph{IEEE Transactions on Neural Networks and Learning Systems},
  vol.~35, no.~12, pp. 17\,202--17\,215, 2024.

\bibitem{XiaShuyin2019INS}
S.~Xia, Y.~Liu, X.~Ding, G.~Wang, H.~Yu, and Y.~Luo, ``Granular ball computing
  classifiers for efficient, scalable and robust learning,'' \emph{Information
  Sciences}, vol. 483, pp. 136--152, 2019.

\bibitem{XieJiang2023TKDE}
J.~Xie, W.~Kong, S.~Xia, G.~Wang, and X.~Gao, ``An efficient spectral
  clustering algorithm based on granular-ball,'' \emph{IEEE Transactions on
  Knowledge and Data Engineering}, vol.~35, no.~9, pp. 9743--9753, 2023.

\bibitem{ChengDongdong2024INS}
D.~Cheng, C.~Zhang, Y.~Li, S.~Xia, G.~Wang, J.~Huang, S.~Zhang, and J.~Xie,
  ``{GB}-{DBSCAN}: {A} fast granular-ball based {DBSCAN} clustering
  algorithm,'' \emph{Information Sciences}, vol. 674, 2024, {A}rt. no. 120731.

\bibitem{XiaShuyin2024TNNLS}
\BIBentryALTinterwordspacing
S.~Xia, B.~Shi, Y.~Wang, J.~Xie, G.~Wang, and X.~Gao, ``{GBCT}: Efficient and
  adaptive clustering via granular-ball computing for complex data,''
  \emph{IEEE Transactions on Neural Networks and Learning Systems}, pp. 1--14,
  2024. [Online]. Available: \url{http://doi.org/10.1109/TNNLS.2024.3497174}
\BIBentrySTDinterwordspacing

\bibitem{XieJiang2024ICDE}
J.~Xie, M.~Dai, S.~Xia, J.~Zhang, G.~Wang, and X.~Gao, ``An efficient fuzzy
  stream clustering method based on granular-ball structure,'' in
  \emph{Proceedings of the {IEEE} 40th {International} {Conference} on {Data}
  {Engineering} ({ICDE})}, 2024, pp. 901--913.

\bibitem{XieJiang2024ICDE2}
J.~Xie, C.~Hua, S.~Xia, Y.~Cheng, G.~Wang, and X.~Gao, ``{W-GBC}: {An} adaptive
  weighted clustering method based on granular-ball structure,'' in
  \emph{Proceedings of the {IEEE} 40th {International} {Conference} on {Data}
  {Engineering} ({ICDE})}, 2024, pp. 914--925.

\bibitem{JiaZihang2025TCYB}
Z.~Jia, Z.~Zhang, and W.~Pedrycz, ``Generation of granular-balls for clustering
  based on the principle of justifiable granularity,'' \emph{IEEE Transactions
  on Cybernetics}, vol.~55, no.~4, pp. 1687--1700, 2025.

\bibitem{XieJiang2024TPAMI}
J.~Xie, X.~Xiang, S.~Xia, L.~Jiang, G.~Wang, and X.~Gao, ``{MGNR}: A
  multi-granularity neighbor relationship and its application in {KNN}
  classification and clustering methods,'' \emph{IEEE Transactions on Pattern
  Analysis and Machine Intelligence}, vol.~46, no.~12, pp. 7956--7972, 2024.

\bibitem{XieQin2024IEEETETCI}
Q.~Xie, Q.~Zhang, S.~Xia, F.~Zhao, C.~Wu, G.~Wang, and W.~Ding, ``{GBG}++: A
  fast and stable granular ball generation method for classification,''
  \emph{IEEE Transactions on Emerging Topics in Computational Intelligence},
  vol.~8, no.~2, pp. 2022--2036, 2024.

\bibitem{XiaShuyin2024TNNLS2}
S.~Xia, X.~Dai, G.~Wang, X.~Gao, and E.~Giem, ``An efficient and adaptive
  granular-ball generation method in classification problem,'' \emph{IEEE
  Transactions on Neural Networks and Learning Systems}, vol.~35, no.~4, pp.
  5319--5331, 2024.

\bibitem{Sajid2025PR}
M.~Sajid, A.~Quadir, and M.~Tanveer, ``{GB-RVFL}: Fusion of randomized neural
  network and granular ball computing,'' \emph{Pattern Recognition}, vol. 159,
  2025, {A}rt. no. 111142.

\bibitem{YangJie2024IEEETFS}
J.~Yang, Z.~Liu, S.~Xia, G.~Wang, Q.~Zhang, S.~Li, and T.~Xu, ``{3WC-GBNRS}++:
  A novel three-way classifier with granular-ball neighborhood rough sets based
  on uncertainty,'' \emph{IEEE Transactions on Fuzzy Systems}, vol.~32, no.~8,
  pp. 4376--4387, 2024.

\bibitem{SunLin2024TFS}
L.~Sun, H.~Liang, W.~Ding, and J.~Xu, ``Granular ball fuzzy neighborhood rough
  sets-based feature selection via multiobjective mayfly optimization,''
  \emph{IEEE Transactions on Fuzzy Systems}, vol.~32, no.~11, pp. 6112--6124,
  2024.

\bibitem{QianWenbin2024TKDE}
W.~Qian, Y.~Li, Q.~Ye, S.~Xia, J.~Huang, and W.~Ding, ``Confidence-induced
  granular partial label feature selection via dependency and similarity,''
  \emph{IEEE Transactions on Knowledge and Data Engineering}, vol.~36, no.~11,
  pp. 5797--5810, 2024.

\bibitem{XiaShuyin2023TNNLS}
S.~Xia, S.~Zheng, G.~Wang, X.~Gao, and B.~Wang, ``Granular ball sampling for
  noisy label classification or imbalanced classification,'' \emph{IEEE
  Transactions on Neural Networks and Learning Systems}, vol.~34, no.~4, pp.
  2144--2155, 2023.

\bibitem{GaoCan2025TKDE}
C.~Gao, X.~Tan, J.~Zhou, W.~Ding, and W.~Pedrycz, ``Fuzzy granule density-based
  outlier detection with multi-scale granular balls,'' \emph{IEEE Transactions
  on Knowledge and Data Engineering}, vol.~37, no.~3, pp. 1182--1197, 2025.

\bibitem{ChengShitong2024PR}
S.~Cheng, X.~Su, B.~Chen, H.~Chen, D.~Peng, and Z.~Yuan, ``{GBMOD}: {A}
  granular-ball mean-shift outlier detector,'' \emph{Pattern Recognition},
  2024, {A}rt. no. 111115.

\bibitem{Pedrycz2013ASOCO}
W.~Pedrycz and W.~Homenda, ``Building the fundamentals of granular computing: A
  principle of justifiable granularity,'' \emph{Applied Soft Computing},
  vol.~13, no.~10, pp. 4209--4218, 2013.

\bibitem{Pedrycz2024TCYB}
W.~Pedrycz, ``Granular computing for machine learning: Pursuing new development
  horizons,'' \emph{IEEE Transactions on Cybernetics}, vol.~55, no.~1, pp.
  460--471, 2025.

\bibitem{Lande1977SZ}
R.~Lande, ``On comparing coefficients of variation,'' \emph{Systematic
  Zoology}, vol.~26, no.~2, pp. 214--217, 1977.

\bibitem{Ross2014IPSES}
S.~M. Ross, \emph{Introduction to Probability and Statistics for Engineers and
  Scientists}, 5th~ed.\hskip 1em plus 0.5em minus 0.4em\relax San Diego, CA,
  USA: Elsevier, 2014.

\bibitem{WangGuoyin2017GrC}
G.~Wang, ``{DGCC}: {Data}-driven granular cognitive computing,'' \emph{Granular
  Computing}, vol.~2, no.~4, pp. 343--355, 2017.

\bibitem{Bache2017UCIDatasets}
\BIBentryALTinterwordspacing
K.~Bache and M.~Lichman, ``{UCI} machine learning repository,'' 2017,
  {D}atasets. [Online]. Available: \url{http://archive.ics.uci.edu/ml}
\BIBentrySTDinterwordspacing

\bibitem{HuangDong2020TKDE}
D.~Huang, C.~Wang, J.~Wu, J.~Lai, and C.~Kwoh, ``Ultra-scalable spectral
  clustering and ensemble clustering,'' \emph{IEEE Transactions on Knowledge
  and Data Engineering}, vol.~32, no.~6, pp. 1212--1226, 2020.

\bibitem{Seyedi2019ESWA}
S.~A. Seyedi, A.~Lotfi, P.~Moradi, and N.~N. Qader, ``Dynamic graph-based label
  propagation for density peaks clustering,'' \emph{Expert Systems with
  Applications}, vol. 115, pp. 314--328, 2019.

\bibitem{GuanJunyi2021Neurocomputing}
J.~Guan, S.~Li, X.~He, J.~Zhu, and J.~Chen, ``Fast hierarchical clustering of
  local density peaks via an association degree transfer method,''
  \emph{Neurocomputing}, vol. 455, pp. 401--418, 2021.

\bibitem{Sander1998DMKD}
J.~Sander, M.~Ester, H.-P. Kriegel, and X.~Xu, ``Density-based clustering in
  spatial databases: {The} algorithm {GDBSCAN} and its applications,''
  \emph{Data Mining and Knowledge Discovery}, vol.~2, no.~2, pp. 169--194,
  1998.

\bibitem{Romano2014ICML}
S.~Romano, J.~Bailey, N.~X. Vinh, and K.~Verspoor, ``Standardized mutual
  information for clustering comparisons: One step further in adjustment for
  chance,'' in \emph{Proceedings of the 31st International Conference on
  Machine Learning (ICML)}, vol.~4, 2014, pp. 2873--2882.

\bibitem{Friedman1940ACO}
M.~Friedman, ``A comparison of alternative tests of significance for the
  problem of $m$ rankings,'' \emph{Annals of Mathematical Statistics}, vol.~11,
  pp. 86--92, 1940.

\bibitem{Dunn1961MultipleCA}
O.~J. Dunn, ``Multiple comparisons among means,'' \emph{Journal of the American
  Statistical Association}, vol.~56, pp. 52--64, 1961.

\bibitem{ZhuQingsheng2016PRL}
Q.~Zhu, J.~Feng, and J.~Huang, ``Natural neighbor: {A} self-adaptive
  neighborhood method without parameter {K},'' \emph{Pattern Recognition
  Letters}, vol.~80, pp. 30--36, 2016.

\end{thebibliography}
\bibliographystyle{IEEEtran}
% argument is your BibTeX string definitions and bibliography database(s)

% Generated by IEEEtran.bst, version: 1.13 (2008/09/30)

% <OR> manually copy in the resultant .bbl file
% set second argument of \begin to the number of references
% (used to reserve space for the reference number labels box)
%\begin{thebibliography}{1}

%\bibitem{IEEEhowto:kopka}
%H.~Kopka and P.~W. Daly, \emph{A Guide to \LaTeX}, 3rd~ed.\hskip 1em plus
%  0.5em minus 0.4em\relax Harlow, England: Addison-Wesley, 1999.

%\end{thebibliography}

% biography section
%
% If you have an EPS/PDF photo (graphicx package needed) extra braces are
% needed around the contents of the optional argument to biography to prevent
% the LaTeX parser from getting confused when it sees the complicated
% \includegraphics command within an optional argument. (You could create
% your own custom macro containing the \includegraphics command to make things
% simpler here.)
%\begin{IEEEbiography}[{\includegraphics[width=1in,height=1.25in,clip,keepaspectratio]{mshell}}]{Michael Shell}
% or if you just want to reserve a space for a photo:

\end{document}